\documentclass{article}

\usepackage[preprint, nonatbib]{neurips_2022}

\usepackage[utf8]{inputenc} 
\usepackage[T1]{fontenc}    
\usepackage{hyperref}       
\usepackage{url}            
\usepackage{booktabs}       
\usepackage{amsfonts}       
\usepackage{nicefrac}       
\usepackage{microtype}      
\usepackage{xcolor}         

\usepackage[
    natbib=true,
    citestyle=numeric-comp,
    bibstyle=ieee,
    sorting=nty
]{biblatex}
\usepackage{graphicx}
\usepackage{caption}
\usepackage{subcaption}
\usepackage{subcaption}
\usepackage{bm}
\usepackage{amsmath,amssymb}
\usepackage{booktabs}
\usepackage{algorithm}
\usepackage[noend]{algorithmic}
\usepackage{multirow}
\usepackage[flushleft]{threeparttable}

\usepackage{multirow}
\usepackage{makecell}
\usepackage{multirow}
\usepackage{xspace}
\usepackage{color}

\newcommand{\ie}{\emph{i.e.,}\xspace}
\newcommand{\eg}{\emph{e.g.,}\xspace}

\title{Sequence Parallelism: Long Sequence Training from System Perspective}

\stepcounter{footnote}

\author{Anonymous Author(s) \\ 
Address}

\author{Shenggui Li$^{1,}$ \quad Fuzhao Xue$^{1}$\thanks{Equal Contribution}  \quad Chaitanya Baranwal$^1$ \quad Yongbin Li$^1$ \quad Yang You$^1$\\
$^1$Department of Computer Science, National University of Singapore
}

\begin{document}

\maketitle

\begin{abstract}

Transformer achieves promising results on various tasks. However, self-attention suffers from quadratic memory requirements with respect to the sequence length. Existing work focuses on reducing time and space complexity from an algorithm perspective. In this work, we propose sequence parallelism, a memory-efficient parallelism method to help us break input sequence length limitation and train with longer sequences on GPUs efficiently. Our approach is compatible with most existing parallelisms (\eg data parallelism, pipeline parallelism and tensor parallelism), which means our sequence parallelism makes 4D parallelism possible. More importantly, we no longer require a single device to hold the whole sequence. That is, with sparse attention, our sequence parallelism enables us to train transformer with infinite long sequence. Specifically, we split the input sequence into multiple chunks and feed each chunk into its corresponding device (\ie GPU). To compute the attention output, we integrated ring-style communication with self-attention calculation and proposed Ring Self-Attention (RSA). Experiments show that sequence parallelism performs well when scaling with batch size and sequence length. Compared with tensor parallelism, our approach achieved $13.7\times$ and $3.0\times$ maximum batch size and sequence length respectively when scaling up to 64 NVIDIA P100 GPUs. With sparse attention, sequence can handle sequence with over 114K tokens, which is over $27\times$ longer than existing sparse attention works holding the whole sequence on a single device.
\end{abstract}

\section{Introduction}\label{Introduction}

Transformer-based language models~\citep{radford2019language,brown2020language,devlin2018bert} have achieved impressive performance on various natural language understanding and generation tasks (\eg Q\&A~\citep{qu2019bert,yang2020bert}, relation extraction~\citep{xue2020gdpnet,xue2020embarrassingly,zhou2020document} and dialogue system~\citep{ni2021recent}). Recently, Transformer also achieved promising results on computer vision tasks~\citep{dosovitskiy2020image,zhang2020span,zhang2021natural} and even on bioinformatics tasks~\citep{elnaggar2020prottrans,wang2021pssm}. These Transformer-based models learn powerful context-aware representation by applying self-attention to all pairs of tokens from the input sequence. This mechanism captures long-term dependencies at the token level for sequence modeling. However, self-attention suffers from quadratic memory requirements with respect to sequence length. Existing works focusing on long sequence modeling devote to solve this problem from algorithm perspective. That is, these works mainly try to reduce the time and space complexity of attention. In this paper, we focus on solving the long sequence training problem from system perspective. Existing system requires us to hold the whole sequence in one GPU, which limits the length of input sequence. Unfortunately, the long sequence is common in real-world applications. For instance, when we train Transformer for medical image classification, each image is much larger than it is in usual (\eg 512$\times$512$\times$512 vs 256$\times$256$\times$3). Then, each medical image has much more tokens (\ie over $512\times$). Each input sequence is much longer than usual. In this case, it is challenging to hold the whole sequence within single GPU.

In this paper, we designed and implemented sequence parallelism, which aims at breaking the limitation that we must store the whole sequence in one GPU. The proposed system can train transformer-based models with longer sequences and a larger batch size. Specifically, we first split the input sequence into multiple chunks along the sequence dimension and feed each sub-sequence chunk to one corresponding GPU. Each GPU thus only holds a part of the full sequence, \ie a sub-sequence. To apply self-attention to the tokens from different chunks, the main challenge is to compute attention scores and outputs across GPUs efficiently. To tackle this problem, we proposed Ring Self-Attention (RSA), which circulates key and value embeddings across GPUs in a ring manner. In this case, each device is just required to keep the attention embeddings corresponding to its own sub-sequence. As a result, our sequence parallelism is memory-efficient, especially for long input sequences.


To model long sequences, existing works mainly focus on sparse attention (\eg \cite{zaheer2020bigbird}) with linear instead of quadratic space complexity. In this paper, we aim to solve the long sequence modeling problem from the distributed system perspective. Compared with sparse attention, we devote ourselves to designing and implementing a system instead of a deep learning algorithm to train attention-based models with longer sequences.We also evaluated our system on sparse attention setting. Existing pipeline parallelism~\cite{huang2018gpipe} and tensor parallelism~\cite{shoeybi2019megatron}) are designed to cope with a larger model size instead of longer sequences. However, when the sequence is long, the challenge is, existing parallelism must keep the whole sequence on one single device. Even if splitting model along hidden and attention-head dimension (\ie tensor parallelism) or depth dimension (\ie pipeline parallelism) can still process longer sequences to some extent, the attention-head and depth are much smaller than sequence length (\eg 12 vs 512), which limits the training scalability and the maximum length of the input sequence. In contrast, our approach splits the whole sequence into multiple devices, enabling it to fit longer input data.




In summary, our main contributions are three folds:
\begin{itemize}

\item  Our system breaks the length limitation of Transformer model training. Sequence parallelism splits long sequences into multiple chunks and feeds them into different devices. It is memory-efficient because each device only keeps the attention embeddings corresponding to its own sub-sequences. With linear space complexity attention, sequence parallelism can help us train the attention model with infinite long sequences. 

\item  To our best knowledge, our work first proposed to use distributed system to handle long sequence training for attention-based models. Our implementation is fully based on PyTorch and is compatible with data parallelism, pipeline parallelism, and tensor parallelism without any extra compiler or library. This makes it possible to integrate sequence parallelism with data parallelism, pipeline parallelism and tensor parallelism into 4D parallelism, and pave the way to train large-scale models with long sequences. 

\item  Our system achieves $3.0\times$ maximum sequence length than SoTA (\ie tensor parallelism) when scaling up to 64 NVIDIA P100 GPUs. On shorter sequence modeling, our system is still more memory-efficient, which achieves $13.7\times$ maximum batch size. With sparse attention, sequence can handle sequence with over 114K tokens, which is over $27\times$ longer than existing sparse attention works holding the whole sequence on a single device. 

\end{itemize}





\section{Background}\label{Background}


\paragraph{Self-attention} We first briefly review the self-attention mechanism in Transformer. For an input sentence $X = \{x_1, \ldots, x_N\}$ with $N$ tokens, we encode every token $x$ into three attention embeddings (\ie query $q$, key $k$, value $v$). To model the dependency among tokens, self-attention computes the attention scores for each token $x_i$ against all other tokens in $X$ by multiplying $q_i$ with $k$ of all tokens. For parallel computing, $q$, $k$ and $v$ of all tokens are combined into three matrices: $Q$, $K$ and $V$. The self-attention of an input sentence $X$ is computed by the following formula:  



\begin{equation}
Attention(Q, K, V) = softmax(\frac{QK^T}{\sqrt{d_k}})V
\end{equation}
where $d_k$ is the dimension of the key. For multi-head attention, please see Appendix~\ref{Appendix:attention} for details.

\begin{figure*}[t]
\centering
\subfloat[Pipeline parallelism]{\label{fig:pipeline_parallelism}
\includegraphics[width=0.28\linewidth]{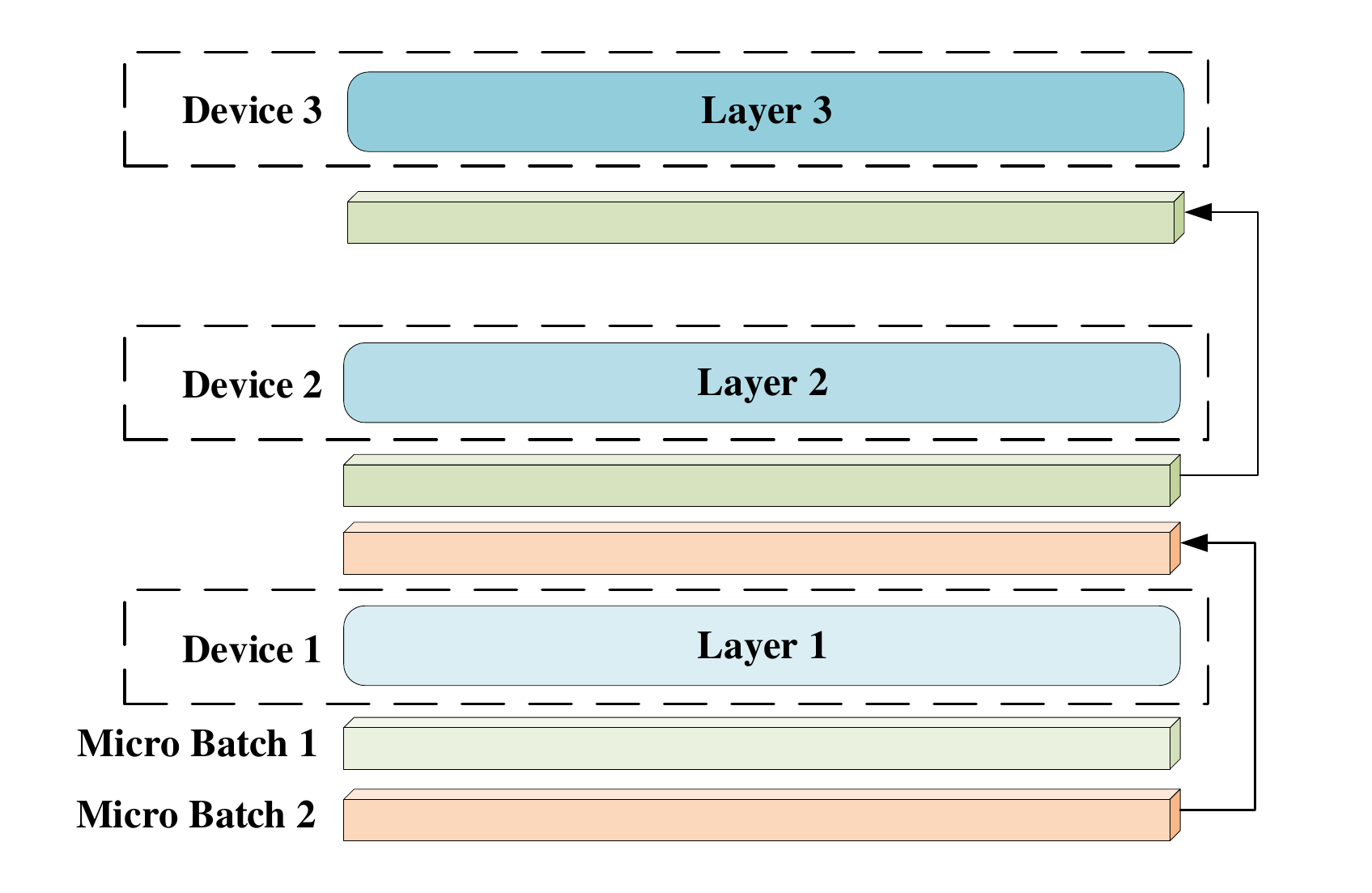}} \hspace{0.2cm}
\subfloat[Tensor parallelism]{\label{fig:tensor_parallelism}
\includegraphics[width=0.28\linewidth]{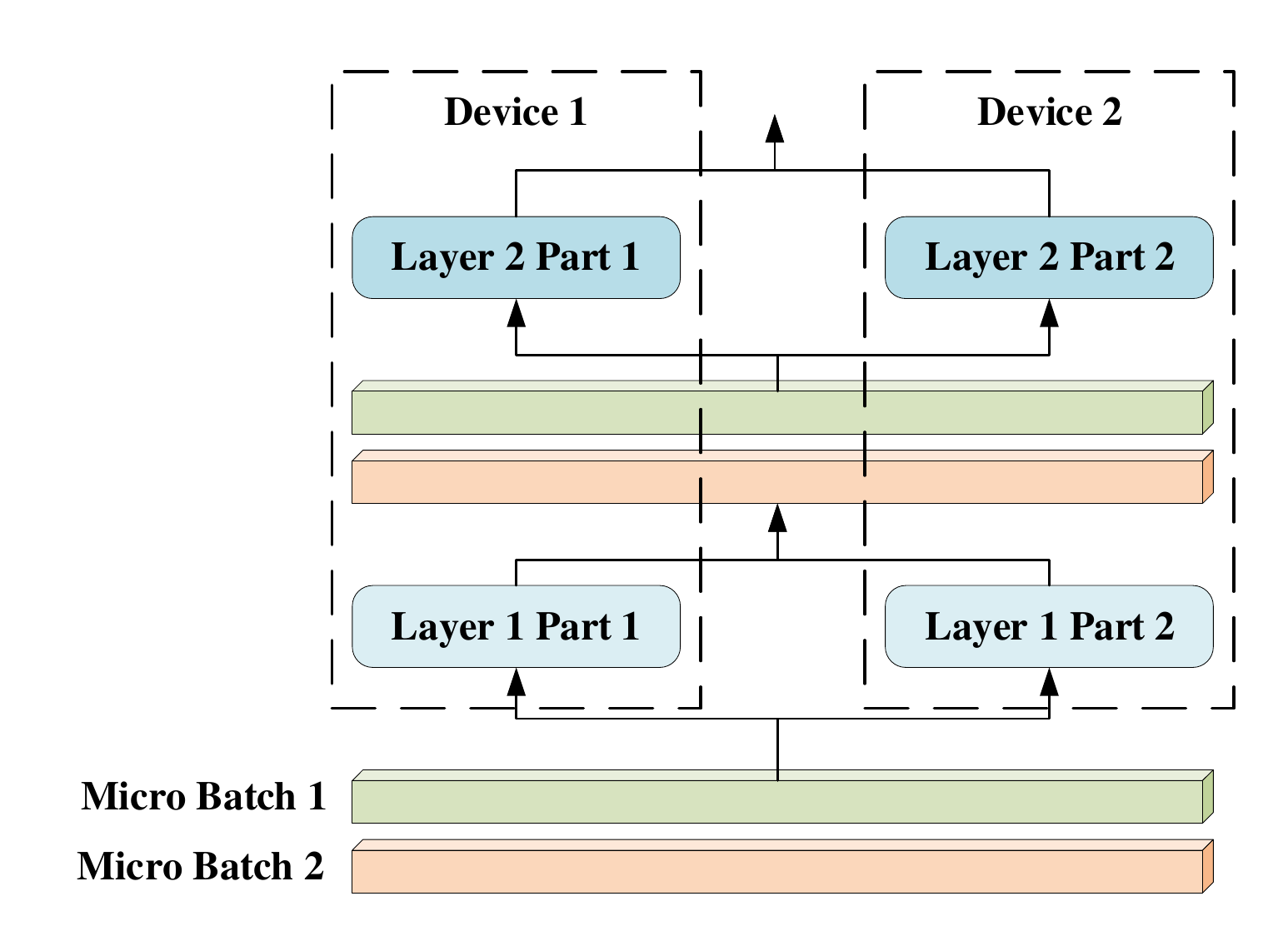}} \hspace{0.2cm}
\subfloat[Sequence parallelism (Ours)]{\label{fig:sequence_parallel}
\includegraphics[width=0.3\textwidth]{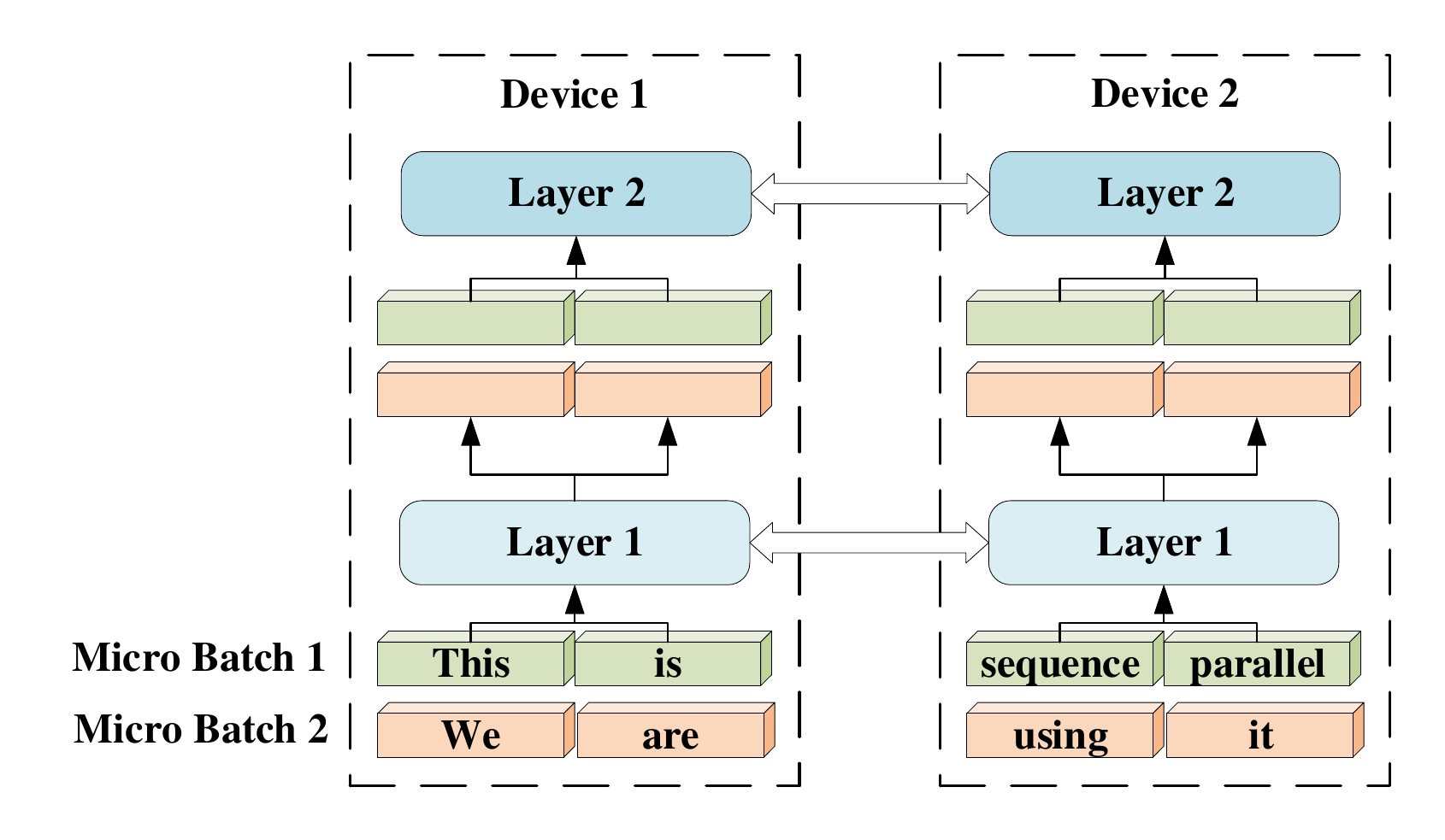}}
\centering
\caption{The overall architecture of the proposed sequence parallelism and existing parallel approaches. For sequence parallelism, Device 1 and Device 2 share the same trainable parameters.}
\label{}
\vspace{-0.6cm}
\end{figure*}

\paragraph{Pipeline parallelism} Huge deep neural networks~\cite{fedus2021switch,2020t5} have shown their effectiveness on various tasks. However, it is challenging to hold the whole model on one single device due to memory limitations. To overcome this, \cite{huang2018gpipe} proposed pipeline parallelism, model parallelism splitting the model layers into different partitions on separate accelerators. As shown in Figure~\ref{fig:pipeline_parallelism}, they split the data along the batch dimension into micro-batches, and each device can process one micro-batch received from the previous device at a time. When the computation is pipelined across micro-batches, pipelining schemes need to ensure that inputs use consistent weight versions for both forward and backward computation to ensure correct weight update and model convergence~\cite{narayanan2021efficient}.



\paragraph{Tensor parallelism} Different from pipeline parallelism which splits models by layer, tensor parallelism (\ie Megatron)~\cite{shoeybi2019megatron}) introduces tensor splitting, where individual layers of the model are partitioned over multiple devices. Similar to our sequence parallelism, tensor parallelism is also designed for Transformer-based models. Each Transformer layer includes a self-attention block and a two-layer multi-layer perceptron (MLP) block. The MLP block can be formalized as:

\begin{equation}
\small
Y=\mathrm{GeLU}(XA),~~Z=YB
\end{equation}
where $GeLU$ is a non-linearity activation function, $X$ is the input data, $Z$ and $Y$ are the outputs. Tensor parallelism splits the weight matrices $A$ and $B$ along columns and rows respectively. Then, the first and second GEMM in the MLP block above can be written as:

\begin{equation}
\small
\begin{aligned}
& [~A~] =
\left[ \begin{array}{cc}
A_1 & A_2
\end{array} 
\right ] \\ 
&\left[ \begin{array}{cc}
Y_1 & Y_2
\end{array} 
\right ] = 
\left[ \begin{array}{cc}
\mathrm{GeLU}(XA_1) & \mathrm{GeLU}(XA_2)
\end{array} 
\right ]\\
& [~B~] =
\left[ \begin{array}{c}
B_1 \\ 
B_2
\end{array} 
\right ] \\ 
& Z = 
\left[ \begin{array}{cc}
Z_1 + Z_2
\end{array} 
\right ] = 
\left[ \begin{array}{cc}
Y_1 & Y_2
\end{array} 
\right ]
\left[ \begin{array}{c}
B_1 \\ 
B_2
\end{array} 
\right ]
\end{aligned}
\end{equation}

At the second GEMM, $Z_1$ and $Z_2$ need to undergo an all-reduce operation to give the final output before the dropout layer in the Transformer layer.


Similarly, Megatron splits the tensors in the self-attention layer as well. For multi-head attention, attention heads are split by column and allocated equally to the devices. The linear layer after the self-attention computation is split by row. An all-reduce operation is needed at the linear layer output to aggregate attention output from all devices. Please refer to Megatron~\cite{shoeybi2019megatron} for more details about tensor parallelism.

\section{Sequence parallelism}\label{Sequence_parallelism}

We propose sequence parallelism for training Transformer with longer sequences. The overview of sequence parallelism is shown in Figure~\ref{fig:sequence_parallel}. Input sequences are split into multiple chunks and the sub-sequences are fed to different corresponding devices. All devices are holding the same trainable parameters but different sub-sequence input chunks. We will introduce and analyze sequence parallelism in detail below. We use the following notation in this section: (1) $\mathrm{B}$: batch size; (2) $\mathrm{L}$: sequence length; (3) $\mathrm{H}$: hidden size of linear layers; (4) $\mathrm{A}$: attention head size; (5) $\mathrm{Z}$: number of attention heads; (6) $\mathrm{N}$: number of GPUs.





\begin{figure*}[t]
\centering
\vspace{-0.6cm}
\subfloat[Transmitting key embeddings among devices to calculate attention scores]{\label{fig:raa_k}
\includegraphics[width=0.7\linewidth]{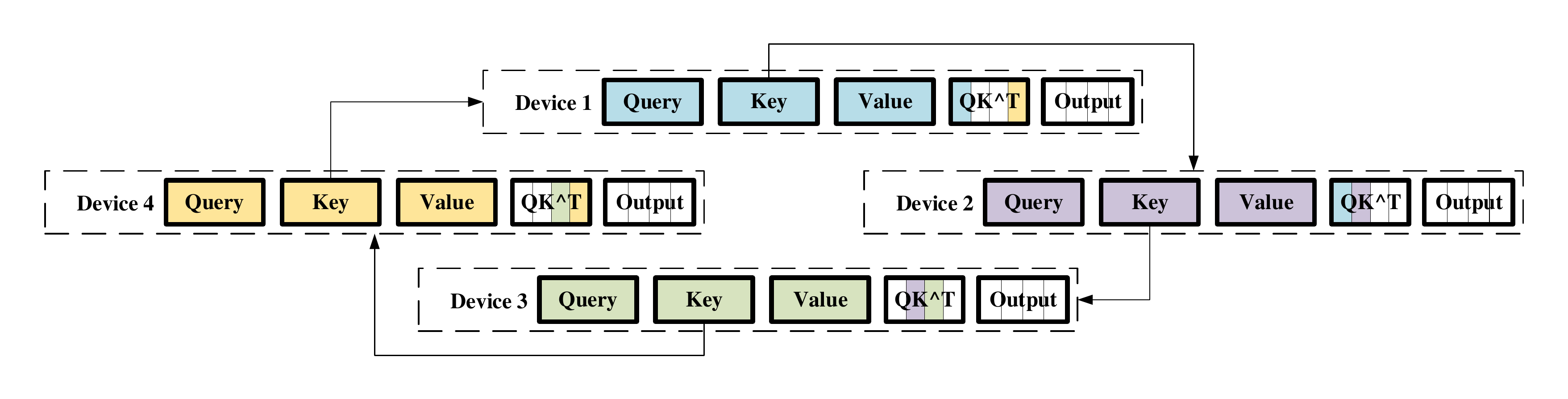}}

\subfloat[Transmitting value embeddings among devices to calculate the output of attention layers]{\label{fig:raa_v}
\includegraphics[width=0.7\linewidth]{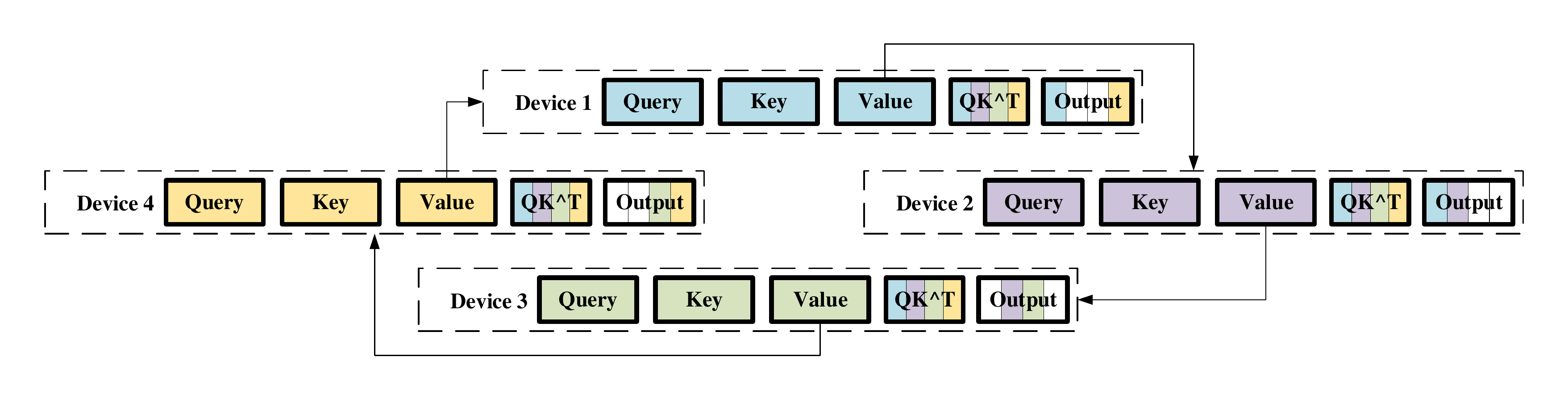}}
\vspace{-0.3cm}
\caption{Ring Self-Attention}
\label{fig:raa}
\vspace{-0.6cm}
\end{figure*}

\subsection{Ring self-Attention}

To distribute sub-sequences to multiple devices, the main challenge is calculating attention scores across devices. Therefore, we propose Ring Self-Attention (RSA) to compute attention output in a distributed setting. There are two steps in RSA to obtain the final output. Please note, we only consider bidirectional self-attention here to introduce RSA succinctly. We treat all heads equally so it can be extended to multi-head attention directly.


Given query embeddings $\{q^1_1,q^1_2,...,q^N_L\}$, key embeddings $\{k^1_1,k^1_2,...,k^N_L\}$ and value embeddings $\{v^1_1,v^1_2,...,v^N_L\}$, where $q^n_s$ represents the key embedding of the $s^{th}$ token in the the sequence which is on $n^{th}$ device. We define all key embeddings on $n^{th}$ device as $K^n$. In RSA, $n^{th}$ device holds the corresponding query embeddings $Q^n$, key embeddings $K^n$ and value embeddings $V^n$. The embeddings on $n^{th}$ device correspond to the $n^{th}$ chunk whose sub-sequence length is $L/N$. Our goal is to obtain $Attention^n(Q^n, K, V)$ which is the self-attention layer output on $n^{th}$ device. To this end, as shown in Figure~\ref{fig:raa_k}, we first transmit the key embeddings among devices to calculate the attention scores $QK^T$ in a circular fashion. Such communication needs to be conducted $N-1$ times to make sure the query embeddings of each sub-sequence can multiply all the key embeddings. To be more specific, each device will compute the partial attention scores based on its local query and key embeddings first. Then, it will receive different key embeddings from the previous device and calculate the partial attention scores with respect to the new key embeddings for each ring-style communication. As a result, all query embeddings $\{Q^1,Q^2,...,Q^N\}$ collected their corresponding attention scores $\{S^1,S^2,...,S^N\}$ on their own devices. 

In the second stage of RSA, we can calculate the self-attention layer output $\{O^1,O^2,...,O^N\}$ based on $\{S^1,S^2,...,S^N\}$ and $\{V^1,V^2,...,V^N\}$. Since computing $O^n$ requires $S^n$ and all value embeddings, as we described in Figure~\ref{fig:raa_v}, we transmit all value embeddings instead of key embeddings in a similar way. For $O^n$, we calculate $S^nV$ by:

\begin{equation}
\small
O^n=S^nV=\sum_{i=1}^N S^n_i V_i
\end{equation}
where $V_i=V^n$, $S^n_i$ is $S^n$ after column splitting, which means $S^n_i \in \mathbb{R}^{L/N \times L/N}$ but $S^n \in \mathbb{R}^{L/N \times L}$.  


\subsection{Modeling}\label{modeling}

We analyzed and compared our sequence parallelism with tensor parallelism in both theoretical modeling and experiments, although tensor parallelism is not our direct baseline. To our best knowledge, sequence parallelism is the first system designed for breaking the length limitation of sequence, so there is actually no direct baseline for sequence parallelism. Therefore, as a distributed training system designed for attention-based models, we compare it with a SoTA model parallelism. Tensor parallelism \cite{narayanan2021efficient} is compatible with data parallelism, pipeline parallelism. Our sequence parallelism is also compatible with them. We expect our system can outperform tensor parallelism with and without pipeline parallelism. We leave integrating sequence parallelism with data parallelism, pipeline parallelism and tensor parallelism into 4D parallelism as our future work. Here, we mainly focus on memory usage and communication cost of tensor parallelism and our sequence parallelism.

\begin{table*}[t]

  \small
  \centering
  \caption{MLP block memory usage comparison. M1 means the matrix before linear layer, and M2 is the trainable matrix of linear layer.}
  \vspace{-2mm}
  \resizebox{0.8\textwidth}{!}{
  \begin{tabular}{llllll}
    \toprule
    & GEMM & M1 & M2 & output  & Memory                                           \\
    \midrule
    \multirow{2}{*}{Tensor parallelism}    & 1st linear & $(\mathrm{B, L, H})$    & $\displaystyle\mathrm{(H, \frac{4H}{N})}$ & $\displaystyle\mathrm{(B, L, \frac{4H}{N})}$  & \multirow{2}{*}{$\displaystyle\mathrm{\frac{32H^2}{N}+\frac{4BLH}{N} + BLH}$}   \\ 
     & 2nd linear & $\displaystyle\mathrm{(B, L, \frac{4H}{N})}$ & $\displaystyle\mathrm{(\frac{4H}{N}, H)}$ & $\displaystyle\mathrm{(B, L, H)}$     &   \\
    \midrule                  
    \multirow{2}{*}{Sequence parallelism}  & 1st linear & $\displaystyle\mathrm{(B, \frac{L}{N}, H)}$    & $\displaystyle\mathrm{(H, 4H)}$ & $\displaystyle\mathrm{(B, \frac{L}{N}, 4H)}$  & \multirow{2}{*}{$\displaystyle\mathrm{32H^2+\frac{5BLH}{N} }$}   \\                
    & 2nd linear &  $\displaystyle\mathrm{(B, \frac{L}{N}, 4H)}$ & $\displaystyle\mathrm{(4H, H)}$ & $\displaystyle\mathrm{(B, \frac{L}{N}, H)}$     &    \\
    \bottomrule
  \end{tabular}
    }
\vspace{-2mm}
\label{MLP_memory_usage_comparison}
\end{table*}

\begin{table*}[t]

\small
  \centering
  \caption{Multi-head attention block memory usage comparison}
  \vspace{-2mm}
  \resizebox{0.8\textwidth}{!}{
  \begin{tabular}{llllll}
    \toprule
    & Operation & M1 & M2 & output  & Memory                                           \\
    \midrule
    \multirow{4}{*}{\makecell[l]{Tensor \\~parallelism}}    & $Q/K/V$ & $(\mathrm{B, L, H})$    & $\displaystyle\mathrm{(H, \frac{ZA}{N})}$ & $\displaystyle\mathrm{(B, \frac{Z}{N}, L, A)}$  &    \\ 
     & $QK^T$ & $\displaystyle\mathrm{(B, \frac{Z}{N}, L, A)}$ & $\displaystyle\mathrm{(B, \frac{Z}{N}, L, A)}$ & $\displaystyle\mathrm{(B, \frac{Z}{N}, L, L)}$     &  $\displaystyle\mathrm{\frac{16AZH}{N} + \frac{4BLZA}{N}}$   \\ 
     & $AV$ & $\displaystyle\mathrm{(B, \frac{Z}{N}, L, L)}$ & $\displaystyle\mathrm{(B, \frac{Z}{N}, L, A)}$ & $\displaystyle\mathrm{(B,\frac{Z}{N}, L, A)}$     & ~~~~~$\displaystyle\mathrm{+ \frac{BZL^2}{N} + BLH}$  \\
     & Linear & $\displaystyle\mathrm{(B, \frac{Z}{N}, L, A)}$ & $\displaystyle\mathrm{(\frac{AZ}{N}, H)}$ & $\displaystyle\mathrm{(B, L, H)}$     &   \\
    \midrule                  
    \multirow{4}{*}{\makecell[l]{Sequence \\~parallelism}}  & $Q/K/V$ & $\displaystyle\mathrm{(B, \frac{L}{N}, H)}$    & $\displaystyle\mathrm{(H, AZ)}$ & $\displaystyle\mathrm{(B, Z, \frac{L}{N}, A)}$  &    \\                
    & Ring-$QK^T$ &  $\displaystyle\mathrm{(B, Z, \frac{L}{N}, A)}$ & $\displaystyle\mathrm{(B, Z, \frac{L}{N}, A)}$ & $\displaystyle\mathrm{(B, Z, \frac{L}{N}, L)}$     & $\displaystyle\mathrm{16AZH + \frac{4BZLA}{N}}$   \\
    & Ring-$AV$ &  $\displaystyle\mathrm{(B, Z, \frac{L}{N}, L)}$ & $\displaystyle\mathrm{(B, Z, \frac{L}{N}, A)}$ & $\displaystyle\mathrm{(B, Z, \frac{L}{N}, A)}$     & ~~~~~$\displaystyle\mathrm{+ \frac{BZL^2}{N} + \frac{BLH}{N}}$   \\
    & Linear &  $\displaystyle\mathrm{(B, Z, \frac{L}{N}, A)}$ & $\displaystyle\mathrm{(AZ, H)}$ & $\displaystyle\mathrm{(B, \frac{L}{N}, H)}$     &    \\

    \bottomrule
  \end{tabular}}
\vspace{-5mm}
\label{Att_memory_usage_comparison}
\end{table*}

\subsubsection{Memory usage}

For memory usage, according to the architecture of Transformer, the comparison is divided into two parts, MLP block and attention block. In this part, we consider multi-head attention instead of self-attention for a fair and accurate comparison. We assume the optimizer is Adam used in Megatron.


\paragraph{MLP block} As shown in Table~\ref{MLP_memory_usage_comparison}, for the MLP blocks, tensor parallelism stores the matrices after row or column-style splitting of the whole sequence. Our sequence parallelism stores the matrices without row or column-style splitting of only one single sub-sequence on each GPU. If we assume that our sequence parallelism is more memory-efficient:

\begin{equation}
\small
\displaystyle\mathrm{\frac{32H^2}{N}+\frac{4BLH}{N} + BLH} > \displaystyle\mathrm{32H^2+\frac{5BLH}{N}}
\end{equation}
We can find that, in MLP blocks, sequence parallelism is more memory-efficient when $\mathrm{BL>32H}$.




\paragraph{Multi-head attention block}  We compared the memory usage of multi-head attention block in Table~\ref{Att_memory_usage_comparison}. Tensor parallelism splits the attention heads here, but our sequence parallelism still splits the length dimension of the sequence data. By comparing the memory usages of multi-head attention block of the two parallelisms, we can find sequence parallelism is more memory-efficient if $\mathrm{BL>16AZ}$. As for communication, tensor parallelism needs an all-reduce operation in both the forward pass and backward pass when calculating the attention output. In our RSA, to facilitate tensor exchange between devices, our communication is equivalent to 2 all-reduce operations in the forward pass and 4 all-reduce operations in the backward pass. The extra communication cost of RSA can be offset by the lack of communication cost in the MLP block.

In both MLP block and multi-head attention block, sequence parallelism is more memory-efficient when we train Transformer with a longer sequence and a larger batch size.

\subsubsection{Communication cost}

Megatron-LM uses all-reduce in its MLP layer and self-attention layer while the communication overhead in sequence parallelism mainly lies in the self-attention layer. Using the same notation as given above, we are able to calculate the amount of data transferred in sequence parallelism and tensor parallelism.

In sequence parallelism, there is no communication in the MLP layer and communication only occurs in the self attention module. There are two ring-style P2P communication in the forward pass for calculating the attention score and attention output respectively. In the backward pass, there are two all-reduce collective communication and two ring-style P2P communication. The amount of data transferred is $2(N-1)*B*Z*(L/N)*A$ in the forward pass and $6(N-1)*B*Z*(L/N)*A$ in the backward pass. The combined amount of data transferred in calculating $QK^T$ and $AV$ will be $8(N-1)*B*Z*(L/N)*A$.

In tensor parallelism of Megatron-LM, the amount of data transferred in the forward pass and backward pass is the same as given by $2(N-1)*B*Z*(L/N)*A$. Since there are 4 collective communication in the forward and backward passes of the MLP layer and self-attention layer, the total communication cost will be $8(N-1)*B*Z*(L/N)*A$.

Thus, sequence parallelism has the same communication overhead compared to tensor parallelism in Megatron-LM. However, please note sequence parallelism has better compatibility with pipeline parallelism, which would reduce the communication overhead above. In tensor parallelism, to save the communication bandwidth between pipeline stages which are often over different nodes, the tensor is split before transmitting to the next stage and all-gathered after transmission. As tensor has already been split along the sequence dimension in sequence parallelism, there is no need to split and all-gather between pipeline stages. Thus, sequence parallelism can have one less all-gather operation per pipeline stage.

%


\section{Experiments}\label{Experiments}

\subsection{Experimental setup}\label{Experiments_setting}
We conducted our experiments on the Piz Daint supercomputer provided by Swiss National Supercomputing Center (CSCS). The Piz Daint supercomputer provides one P100 GPU (16GB GPU RAM) for each compute node and the compute nodes are connected by a high-bandwidth network. We chose two bidirectional language models, namely BERT Base and BERT Large, to evaluate our sequence parallelism. We also verified the convergence performance of sequence parallelism (see Appendix~\ref{Appendix:Convergence}). Since we are using the original model but different systems, the accuracy should be the same. The slight differences are from randomness.

\subsection{Maximum batch size}






\begin{figure}[t]
\centering
\begin{minipage}[b]{0.4\linewidth}
\centering
\includegraphics[width=1.0\textwidth]{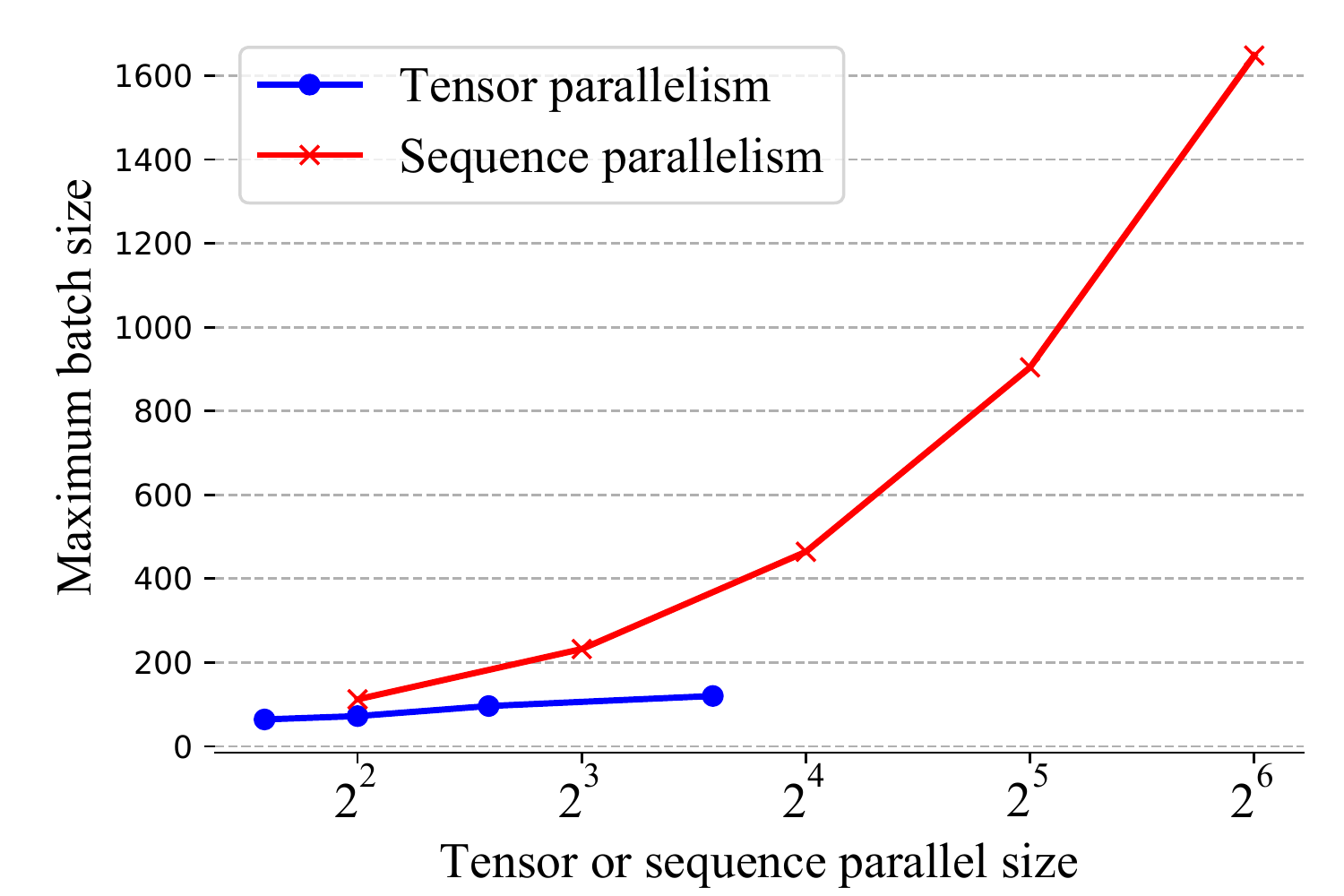}
\subcaption{Maximum batch size of BERT Base scaling along tensor or sequence parallel size}
\label{fig:Batch_BERT_Base_wo_pipeline_batchsize}
\end{minipage} \hspace{1cm}
\begin{minipage}[b]{0.4\linewidth}
\centering
\includegraphics[width=1.0\textwidth]{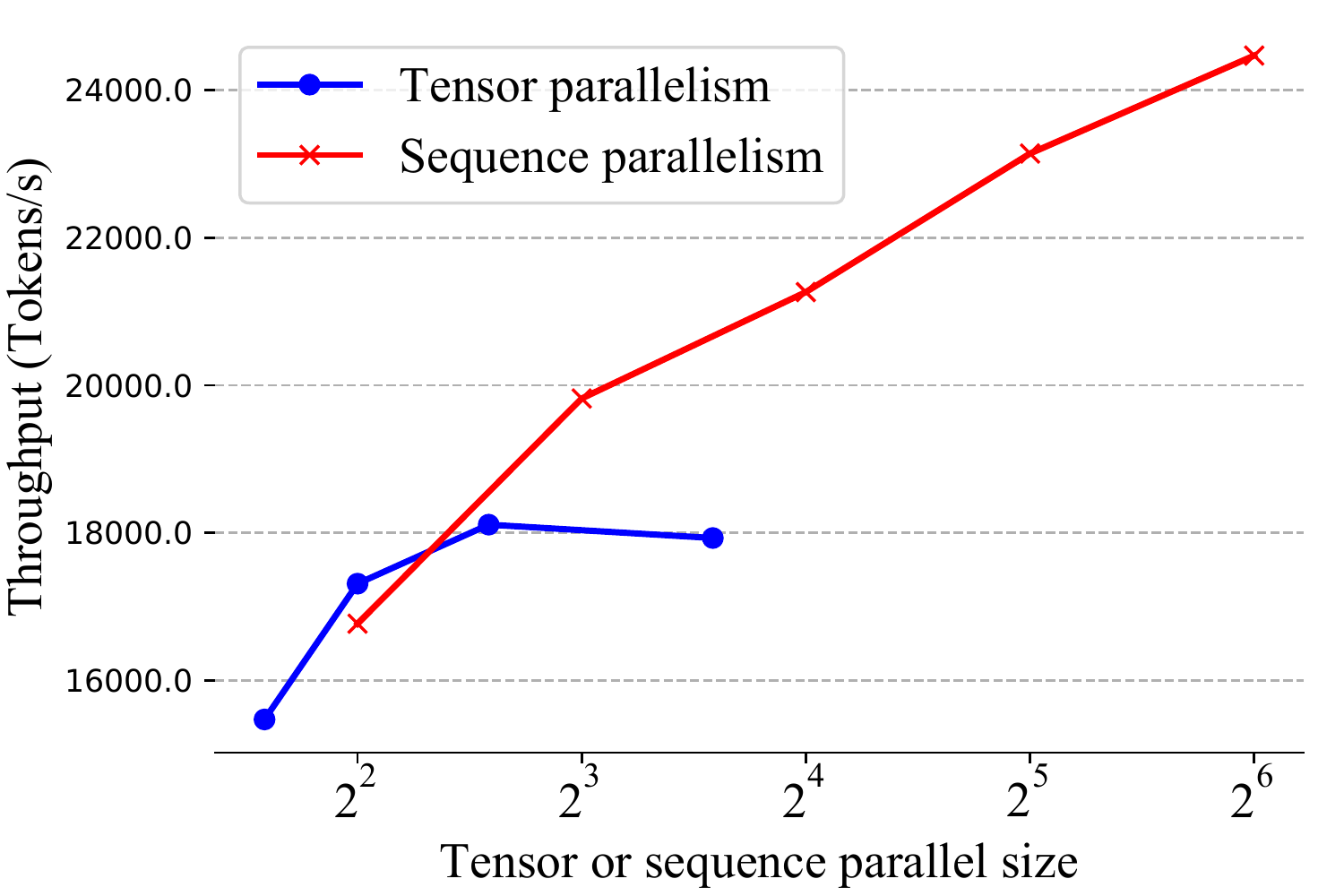}
\subcaption{Throughput of BERT Base scaling along tensor or sequence parallel size}
\label{fig:Batch_BERT_Base_wo_pipeline_throughput}
\end{minipage}
\centering
\caption{Scaling with sequence/tensor parallelism}
\label{fig:maximum batch size strong scaling}
\end{figure}

Since our sequence parallelism is memory-efficient to handle larger batch sizes, we first investigated the maximum batch size we can reach with sequence parallelism. In this section, for a comprehensive comparison, we scaled with tensor or sequence parallelism on BERT Base and BERT Large. We also fixed the tensor or parallel size and then scale them with pipeline parallelism to evaluate the verify the compatibility with pipeline parallelism. We used tokens per second as the metric for throughput. To this end, we trained BERT Base and BERT Large for 150 iterations in total, and then we calculate the mean tokens processed per second within the last 100 iterations.





\paragraph{Scaling with sequence/tensor parallelism} We fixed all hyper-parameters except the batch size and the tensor parallelism or sequence parallelism size. We trained the model with a sequence length of 512 and no pipeline parallelism is used. The tensor parallelism size in Megatron is limited by the number of attention heads and hidden size, because these two hyper-parameters are required to be divisible by the tensor parallelism size. Among them, the number of attention heads is small so it limits the tensor parallelism. Thus, tensor parallelism size is a maximum of 12 for the BERT Base model in Megatron. In contrast, for our sequence parallelism, only the sequence length is required to be divisible by the sequence parallelism size, so that we can scale sequence parallelism to a larger size since it is a much larger hyper-parameter than the number of attention heads. 


For BERT Base, our sequence parallelism outperforms tensor parallelism in terms of memory consumption. Figure~\ref{fig:Batch_BERT_Base_wo_pipeline_batchsize} shows that our system on 64 GPUs can achieve $13.7\times$ larger batch size than Megatron on 12 GPUs. Even if we combine data parallelism and tensor parallelism to scale up to 64 GPUs for Megatron, our system would still support a larger batch size. In Figure~\ref{fig:Batch_BERT_Base_wo_pipeline_throughput}, we can observe sequence parallelism achieved comparable throughput with the same parallel size, and our system can extend to a larger parallel size to achieve better performance. For the results on BERT Large, please see Appendix~\ref{app:scaling_seq_tensor} for details.










\begin{figure}[t]
\centering
\vspace{-0.3cm}
\begin{minipage}[b]{0.4\linewidth}
\centering
\includegraphics[width=1.0\textwidth]{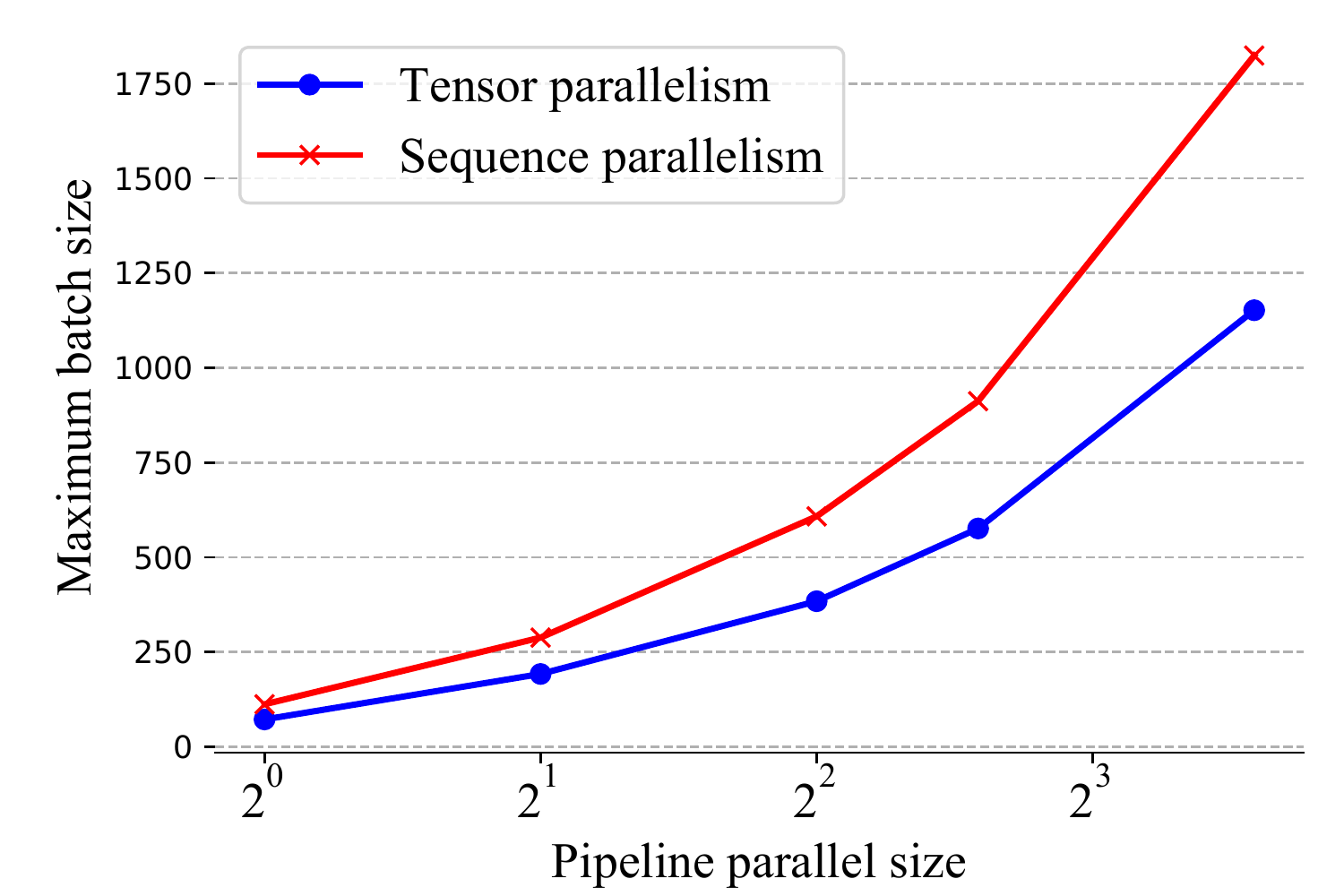}
\subcaption{Maximum batch size of BERT base scaling along pipeline parallel size}
\label{fig:Batch_BERT_Base_w_pipeline_batchsize}
\end{minipage} \hspace{1cm}
\begin{minipage}[b]{0.4\linewidth}
\centering
\includegraphics[width=1.0\textwidth]{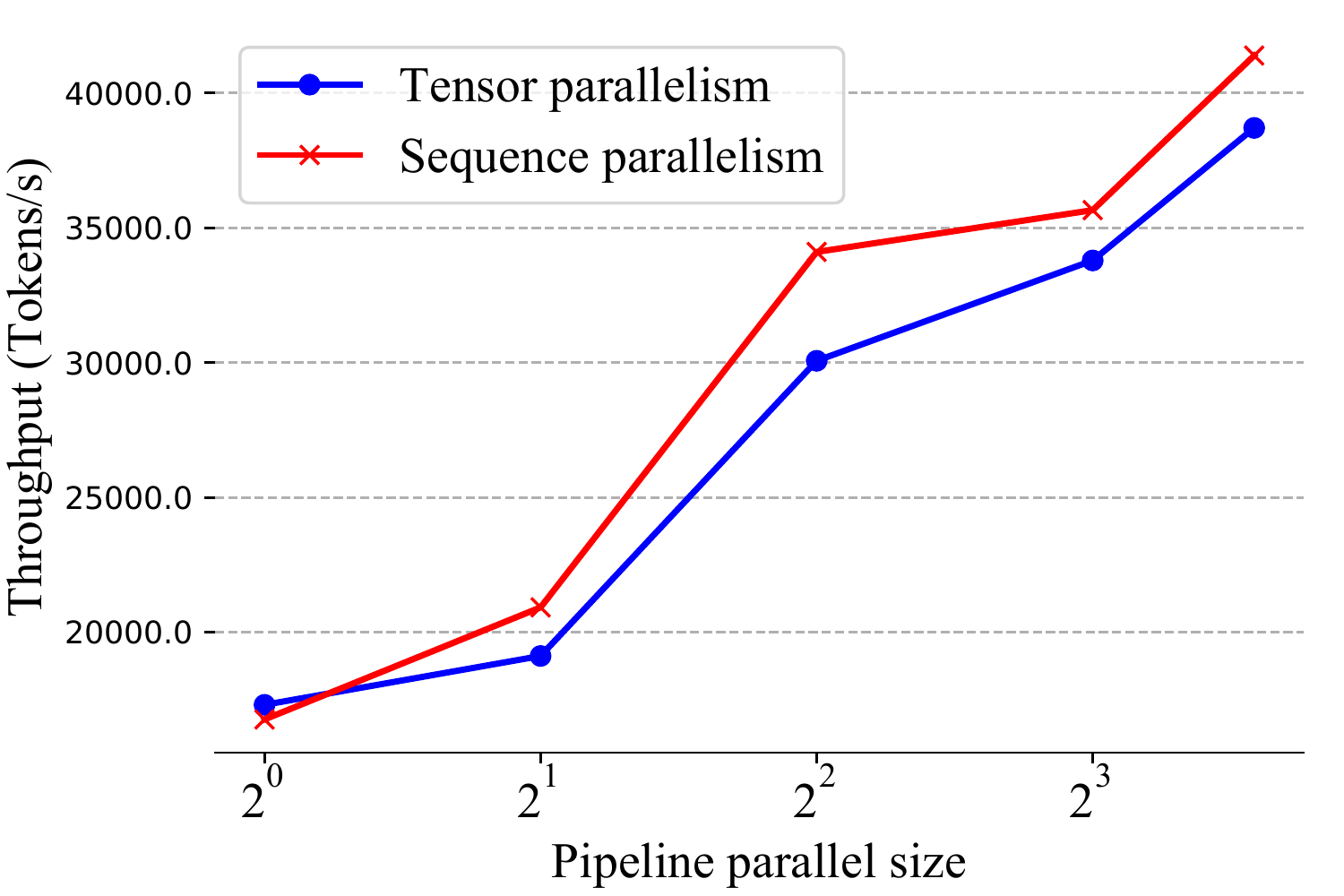}
\subcaption{Throughput of BERT base scaling along pipeline parallel size}
\label{fig:Batch_BERT_Base_w_pipeline_throughput}
\end{minipage}
\centering
\caption{Scaling with pipeline parallelism}
\label{fig:maximum batch size weak scaling}
\vspace{-0.6cm}
\end{figure}


\paragraph{Scaling with pipeline parallelism} To verify the compatibility with pipeline parallelism, we fixed the tensor parallelism and sequence parallelism size as 4 and scale the pipeline parallel size. For BERT Base, we can observe that sequence parallelism outperforms tensor parallelism on the maximum batch size in Figure~\ref{fig:Batch_BERT_Base_w_pipeline_batchsize}. It can be noted that sequence parallelism also achieved higher throughput when using more pipeline stages as shown in Figure~\ref{fig:Batch_BERT_Base_w_pipeline_throughput}. This is because Megatron incurs extra communication costs between pipeline stages. Megatron holds the activation for the full sequence on each device. Thus, it needs to split the activation, transmit the partial activation to the next device, and gather back the partial activation when sending the activation between pipelines. This incurs less communication overhead compared to transmitting the whole activation between pipelines. However, this still brings more communication costs than ours, as no splitting and all-gather operation is required for our sub-sequence intermediate activation. Therefore, our sequence parallelism achieved better throughput when scaling along with pipeline parallel size. 






 





\begin{table*}[t]
\small
  \centering
  \caption{Sparse attention block memory usage. $K$ is the projection dimension in Linformer~\citep{wang2020linformer}}
  \resizebox{0.75\textwidth}{!}{
  \begin{tabular}{llllll}
    \toprule
    & Operation & M1 & M2 & output  & Memory                                           \\ \midrule
    \multirow{5}{*}{\makecell[l]{Linformer\\~Sequence \\~parallelism}}  & $Q/K/V$ & $\displaystyle\mathrm{(B, \frac{L}{N}, H)}$    & $\displaystyle\mathrm{(H, AZ)}$ & $\displaystyle\mathrm{(B, Z, \frac{L}{N}, A)}$  &    \\
    & Projection & $\displaystyle\mathrm{(B, Z, \frac{L}{N}, A)}$    & $\displaystyle\mathrm{(\frac{L}{N}, K)}$ & $\displaystyle\mathrm{(B, Z, K, A)}$  & $\displaystyle\mathrm{2AZH + \frac{2BZLA}{N}}$    \\  
    & Ring-$QK^T$ &  $\displaystyle\mathrm{(B, Z, \frac{L}{N}, A)}$ & $\displaystyle\mathrm{(B, Z, K, A)}$ & $\displaystyle\mathrm{(B, Z, \frac{L}{N}, K)}$     & ~~~~~$\displaystyle\mathrm{+ \frac{BZLK}{N} + \frac{BLH}{N}}$   \\
    & Ring-$AV$ &  $\displaystyle\mathrm{(B, Z, \frac{L}{N}, K)}$ & $\displaystyle\mathrm{(B, Z, K, A)}$ & $\displaystyle\mathrm{(B, Z, \frac{L}{N}, A)}$     &  ~~~~~$\displaystyle\mathrm{+ 2BZKA}$  \\ 
    & Linear &  $\displaystyle\mathrm{(B, Z, \frac{L}{N}, A)}$ & $\displaystyle\mathrm{(AZ, H)}$ & $\displaystyle\mathrm{(B, \frac{L}{N}, H)}$     &    \\

    \bottomrule
  \end{tabular}}
  \vspace{-0.4cm}

\label{tbl:Sparse_att_memory_usage}
\end{table*}

\begin{figure}[t]
\centering
\begin{minipage}[b]{0.4\linewidth}
\centering
\includegraphics[width=1.0\textwidth]{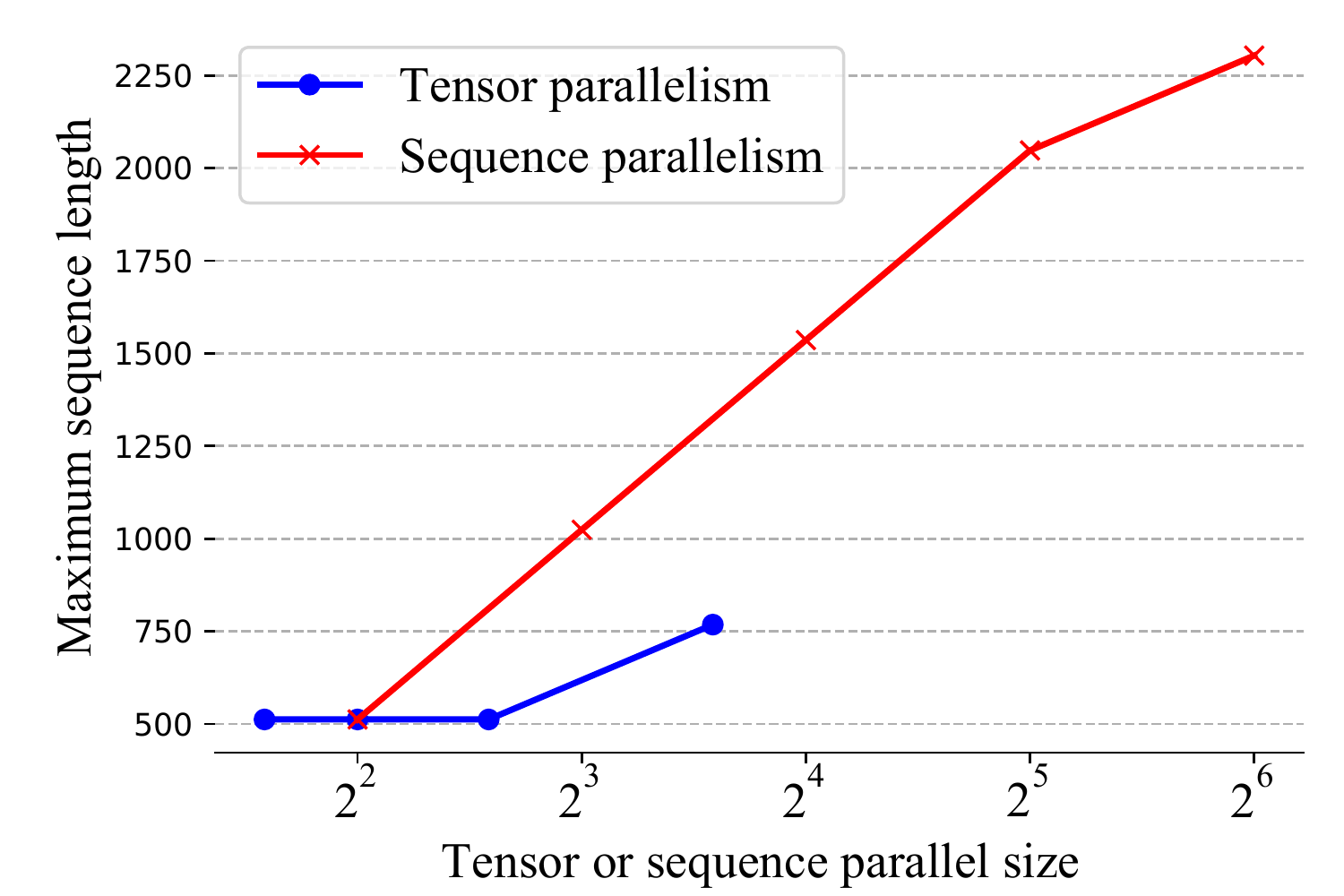}
\subcaption{Maximum sequence length on BERT base}
\label{fig:Length_BERT_Base_wo_pipeline_length}
\end{minipage}\hspace{1cm}
\begin{minipage}[b]{0.4\linewidth}
\centering
\includegraphics[width=1.0\textwidth]{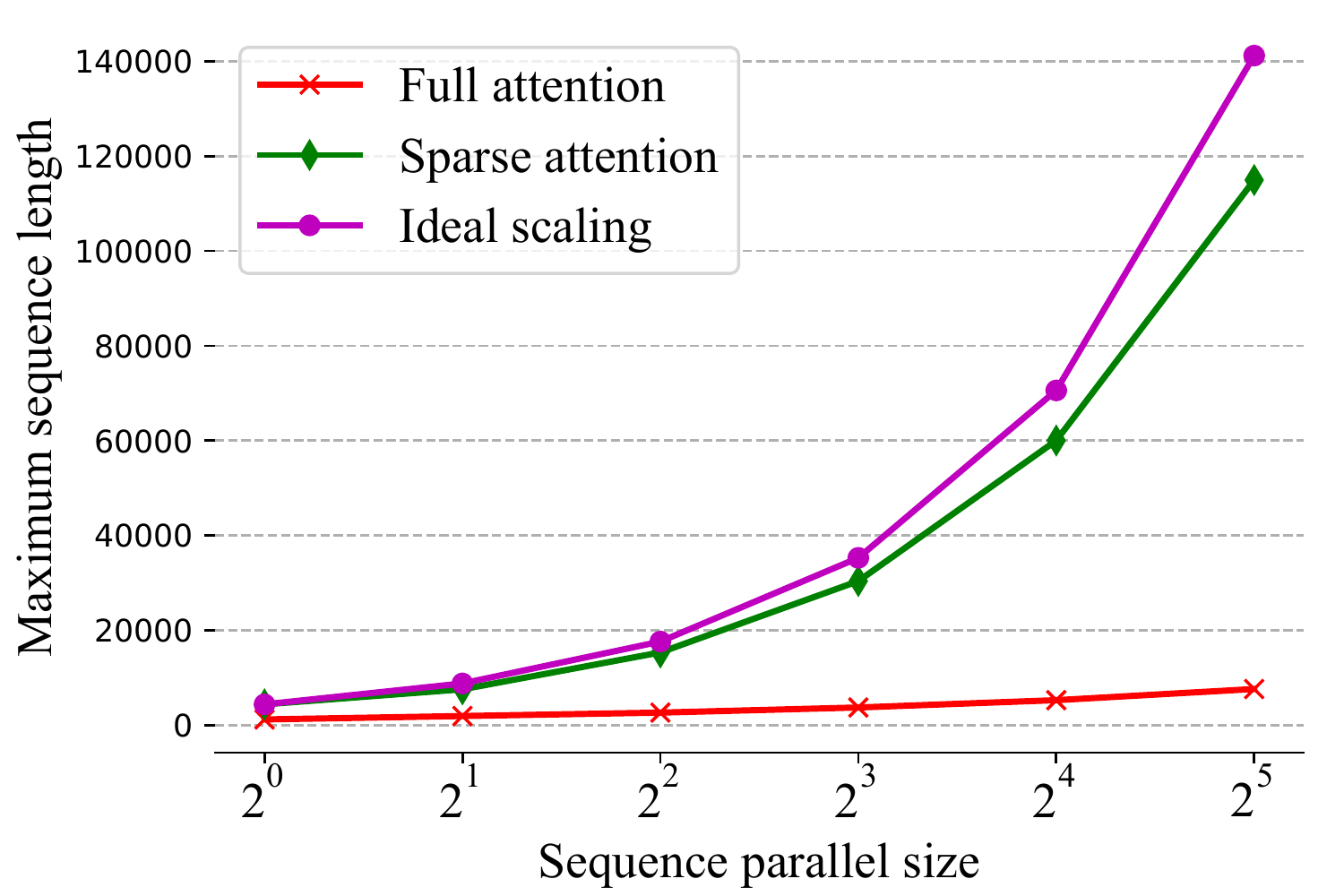}
\subcaption{Sequence length upper bound}
\label{fig:seq_len_upper_bound}
\end{minipage}
\centering

\caption{Scaling with sequence length}
\label{fig:maximum sequence length strong scaling}
\vspace{-0.6cm}
\end{figure}

\subsection{Maximum sequence length}\label{sec:max_length}

Sequence parallelism is designed for training Transformer-based models with longer input sequences, so we investigated the maximum sequence length it can handle. Similarly, we still compared tensor parallelism without pipeline parallelism. 


\paragraph{Compared with tensor parallelism} We fixed batch size as 64 for BERT Base and no pipeline parallelism was used. We show the maximum sequence length in Figure~\ref{fig:Length_BERT_Base_wo_pipeline_length}. If we scale up to 64 GPUs, we can achieve around $3\times$ maximum sequence length on BERT Base. Another observation is splitting along the number of attention heads limits the input sequence length of tensor parallelism in Megatron, but our sequence parallelism can scale easily by splitting a sequence into multiple chunks. When using the same 16 GPUs, our sequence parallelism still can achieve $1.4 \times$ larger sequence length than tensor parallelism. The gap is expected to widen if we use 32GB GPUs instead of 16GB GPUs.





\paragraph{Sequence length upper bound} To investigate the maximum sequence length our system can handle on the cluster with 32 P100 GPUs. we set both data and pipeline parallel size as 1 and global batch size as 4. As sparse attention is widely used in long sequence training, we adapt Linformer~\citep{wang2020linformer}, \ie one sparse attention algorithm with linear time and space complexity. Our sequence parallelism is compatible with the sparse attention. More importantly, as shown in Table~\ref{tbl:Sparse_att_memory_usage}, for memory usage in sparse attention block, all terms including sequence length $L$ is divided by number of devices $N$, which means \textbf{we can scale the sequence length to infinite long if we use sparse attention}. To investigate the sequence length upper bound of sequence length on the sparse attention setting, we compare sequence with sparse and full attention. As shown in Figure~\ref{fig:seq_len_upper_bound}, if we use sparse attention on sequence parallelism, we can almost achieve ideal scaling. With 32 P100 GPUs, our sequence parallelism with sparse attention can handle the sequence with 114K tokens, which is over $27\times$ longer than recent sparse attention papers holding the whole sequence on a single device ~\citep{zaheer2020bigbird, wang2020linformer}.





\begin{table*}[t]
\small
\caption{Weak scaling results. Parallel size is the tensor or sequence parallel size. Batch size denotes global batch size, respectively. Memory and Token/sec denote max allocated memory/MB and tokens processed per second. OOM means that CUDA out of memory occurs. }  
\vspace{-0.2cm}
\centering
\resizebox{0.75\textwidth}{!}{
\begin{tabular}{lll ll ll}
\toprule
\multicolumn{1}{c}{\multirow{2}{*}{Parallel size}} & \multicolumn{1}{c}{\multirow{2}{*}{Batch size}} & \multicolumn{1}{c }{\multirow{2}{*}{Sequence length}} & \multicolumn{2}{c }{Tensor parallelism}    & \multicolumn{2}{c}{Sequence parallelism}        \\ \cmidrule(lr){4-5} \cmidrule(lr){6-7}
\multicolumn{1}{c}{}                                               & \multicolumn{1}{c}{}                                   & \multicolumn{1}{c }{}                                 & Memory & Token/sec & Memory & Token/sec \\ 
\midrule
1                                                                  & 64                                                     & 512                                                  & 8477.28              & 9946.15  & 8477.53              & 9261.04  \\
2                                                                  & 128                                                    & 512                                                  & 9520.47              & 15510.19 & 8478.76              & 13938.22 \\
4                                                                  & 256                                                    & 512                                                  & 12232.52             & 20701.96 & 8481.26              & 21269.91 \\
8                                                                  & 512                                                    & 512                                                  & OOM                  & OOM      & 8490.75              & 26401.64 \\
\midrule
1                                                                  & 64                                                     & 256                                                  & 3707.39              & 9752.61  & 3707.01              & 9340.13  \\
2                                                                  & 64                                                     & 512                                                  & 4993.43              & 14195.17 & 4670.64              & 13144.16 \\
4                                                                  & 64                                                     & 1024                                                 & 8175.93              & 19879.27 & 6601.88              & 18243.82 \\
8                                                                  & 64                                                     & 2048                                                 & 14862.09             & 22330.5  & 10536.38             & 21625.51 \\
\bottomrule
\end{tabular}}

\label{table:Weak scaling}
\vspace{-0.6cm}
\end{table*}

\subsection{Weak scaling}



Strong scaling limits the upper bound of batch size and sequence length within a single device, so we mainly discuss weak scaling in this section. We scale the batch size and sequence length separately when increasing the number of nodes. We fixed the pipeline parallelism size as 8. In Table~\ref{table:Weak scaling}, sequence parallelism achieved almost constant memory usage when scaling along with the global batch size, which outperforms tensor parallelism by a large margin. As for weak scaling along the sequence length, our method still uses much less memory with comparable throughput.








\section{Discussion}\label{Discussion}


Although there are other related works including DeepSpeed~\cite{rasley2020deepspeed}, GShard~\cite{lepikhin2020gshard}, GSPMD~\cite{xu2021gspmd}, etc., they are not our direct baseline in experiments. DeepSpeed is an efficient method to optimize memory footprint in data parallel training by using ZeRO Optimizer~\cite{rajbhandari2021zero} and ZeRO-Offload~\cite{ren2021zerooffload}. DeepSpeed and our method optimize training in different dimensions and they are actually compatible with each other. Our method is orthogonal to DeepSpeed just as how DeepSpeed can be integrated with Megatron. Thus, Megatron should be our baseline.

GShard and GSPMD are two libraries built for the TensorFlow community to partition model parameters in distributed training. GSPMD is developed based on GShard. These two methods rely on the static computation graph of TensorFlow to train larger models while we provide a plug-and-play tool based on PyTorch's dynamic computation graph to train on longer sequences. The difference in the computation paradigms makes them unsuitable as our baseline.

\section{Conclusion}\label{Conclusion}
In this paper, we proposed sequence parallelism for training transformer with longer sequence. Sequence parallelism is designed to break the limitation of sequence length on a single device. We have shown that sequence parallelism can handle longer sequence and is more memory-efficient than SoTA. In particular, sequence parallelism achieves $3.0\times$ maximum sequence length and $13.7\times$ maximum batch size than tensor parallelism when scaling up to 64 GPUs. Unlike both tensor and pipeline parallelism, sequence parallelism is not limited by the smaller hyper-parameters (\eg number of attention heads, number of layers). Therefore, our sequence parallelism can be adapted as long as the sequence length is divisible by sequence parallel size. With sparse attention, sequence parallelism can handle sequence with over 114K tokens, which is over $27\times$ longer than existing sparse attention works holding the whole sequence on a single device. 
We used a language model (\ie BERT) to evaluate our system, but it can also be adapted to vision tasks. This work paves the way to process large images~\cite{hou2019high} by ViT~\cite{dosovitskiy2020image} as a larger image means more patches or longer sequences. In the future, we plan to integrate data, pipeline, tensor and sequence parallelism to construct 4D parallelism. This would enable us to train extremely large models with very long sequences. 

\printbibliography

\newpage

\section*{Checklist}


\begin{enumerate}

\item For all authors...
\begin{enumerate}
  \item Do the main claims made in the abstract and introduction accurately reflect the paper's contributions and scope?
    \answerYes{}
  \item Did you describe the limitations of your work?
    \answerYes{}
  \item Did you discuss any potential negative societal impacts of your work?
    \answerNA{}
  \item Have you read the ethics review guidelines and ensured that your paper conforms to them?
    \answerYes{}
\end{enumerate}

\item If you are including theoretical results...
\begin{enumerate}
  \item Did you state the full set of assumptions of all theoretical results?
    \answerNA{}
  \item Did you include complete proofs of all theoretical results?
    \answerNA{}
\end{enumerate}

\item If you ran experiments...
\begin{enumerate}
  \item Did you include the code, data, and instructions needed to reproduce the main experimental results (either in the supplemental material or as a URL)?
    \answerYes{}
  \item Did you specify all the training details (\eg, data splits, hyperparameters, how they were chosen)?
    \answerYes{}
    \item Did you report error bars (\eg, with respect to the random seed after running experiments multiple times)?
    \answerNA{}
    \item Did you include the total amount of compute and the type of resources used (\eg, type of GPUs, internal cluster, or cloud provider)?
    \answerYes{}
\end{enumerate}

\item If you are using existing assets (\eg, code, data, models) or curating/releasing new assets...
\begin{enumerate}
  \item If your work uses existing assets, did you cite the creators?
    \answerYes{}
  \item Did you mention the license of the assets?
    \answerYes{}
  \item Did you include any new assets either in the supplemental material or as a URL?
    \answerYes{}
  \item Did you discuss whether and how consent was obtained from people whose data you're using/curating?
    \answerNA{}
  \item Did you discuss whether the data you are using/curating contains personally identifiable information or offensive content?
    \answerYes{}
\end{enumerate}

\item If you used crowdsourcing or conducted research with human subjects...
\begin{enumerate}
  \item Did you include the full text of instructions given to participants and screenshots, if applicable?
    \answerNA{}
  \item Did you describe any potential participant risks, with links to Institutional Review Board (IRB) approvals, if applicable?
    \answerNA{}
  \item Did you include the estimated hourly wage paid to participants and the total amount spent on participant compensation?
    \answerNA{}
\end{enumerate}

\end{enumerate}

\newpage
\appendix
\noindent\textbf{\Large Appendix}
\section{Multi-head attnetion}
\label{Appendix:attention}

Multi-head attention is designed to jointly consider the information from different subspaces of embedding. Compared with self-attention below, multi-head attention has $h$ query, key and value embeddings instead of the single one, where $h$ denotes the number of heads. We obtain these embeddings with identical shapes by linear transformations. The multi-head attention can be described as: 

\begin{equation}
MultiHead(Q, K, V) = Concat(head_1, ..., head_h)W^O
\end{equation}
where $head_i  = Attention(Q_i, K_i, V_i)$ and $W$ denotes the linear transformations. All heads are concatenated and further projected by linear transformation $W^O$.

\section{Convergence performance}\label{Appendix:Convergence}

\begin{figure}[h]
\centering
\begin{minipage}[b]{0.45\linewidth}
\centering
\includegraphics[width=1.0\textwidth]{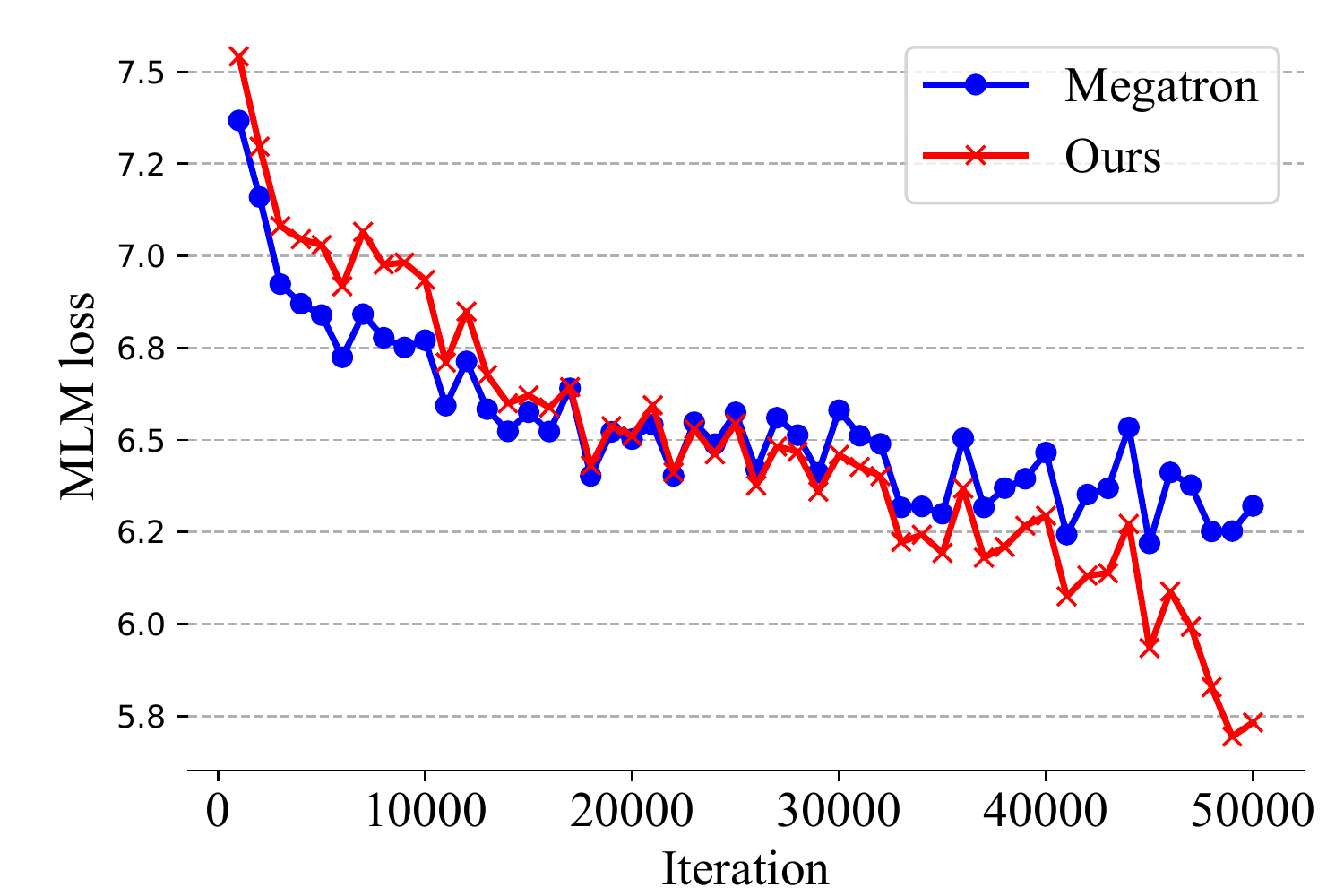}
\subcaption{Convergence performance of MLM loss}
\label{fig:conv_mlm}
\end{minipage}
\begin{minipage}[b]{0.45\linewidth}
\centering
\includegraphics[width=1.0\textwidth]{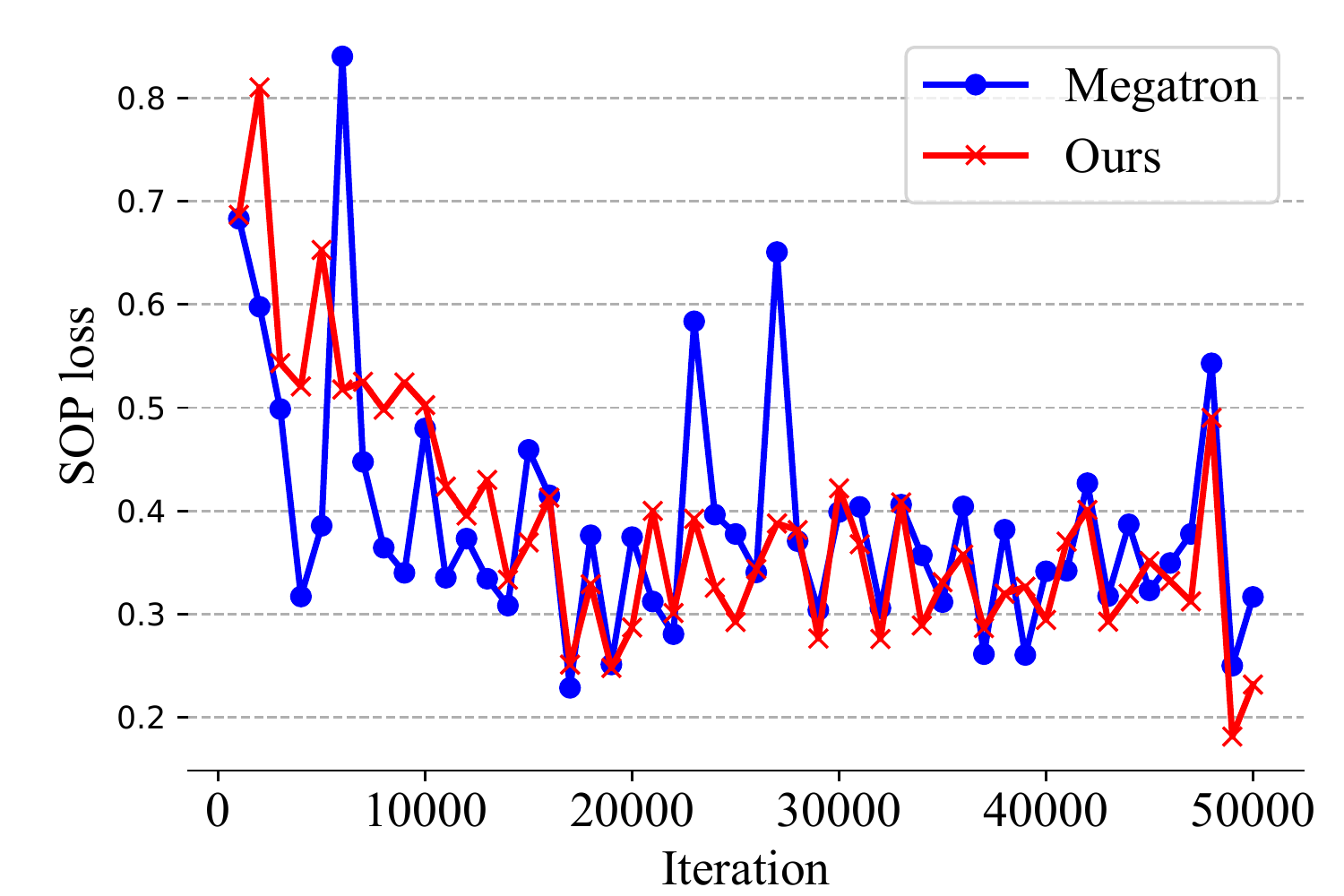}
\subcaption{Convergence performance of SOP loss}
\label{fig:conv_sop}
\end{minipage}
\centering
\caption{Convergence performance comparison between tensor parallelism and ours}
\label{fig:convergence}
\end{figure}

We verified the convergence performance of sequence parallelism. We used the Wikipedia dataset~\cite{devlin2018bert} and evaluated Megatron and our model on the development set every 1k iterations. We trained the BERT Large model for 50k iterations with the default hyper-parameters used by Megatron. Our goal here is to verify the correctness of our implementation so we trained the model for fewer steps. We set parallel size as 4 for tensor parallelism in Megatron and sequence parallelism in our model. No pipeline was used for both models. In Figure~\ref{fig:convergence}, Our sequence parallelism shows good convergence on both the masked language modeling (MLM) loss and the sentence order prediction (SOP) loss. Compared with Megatron, sequence parallelism has a similar trend in convergence and achieved lower values for both MLM loss and SOP loss for 50k iterations.

\section{Scaling with sequence/tensor parallelism}\label{app:scaling_seq_tensor}

\begin{figure}[h]
\centering
\begin{minipage}[b]{0.45\linewidth}
\centering
\includegraphics[width=1.0\textwidth]{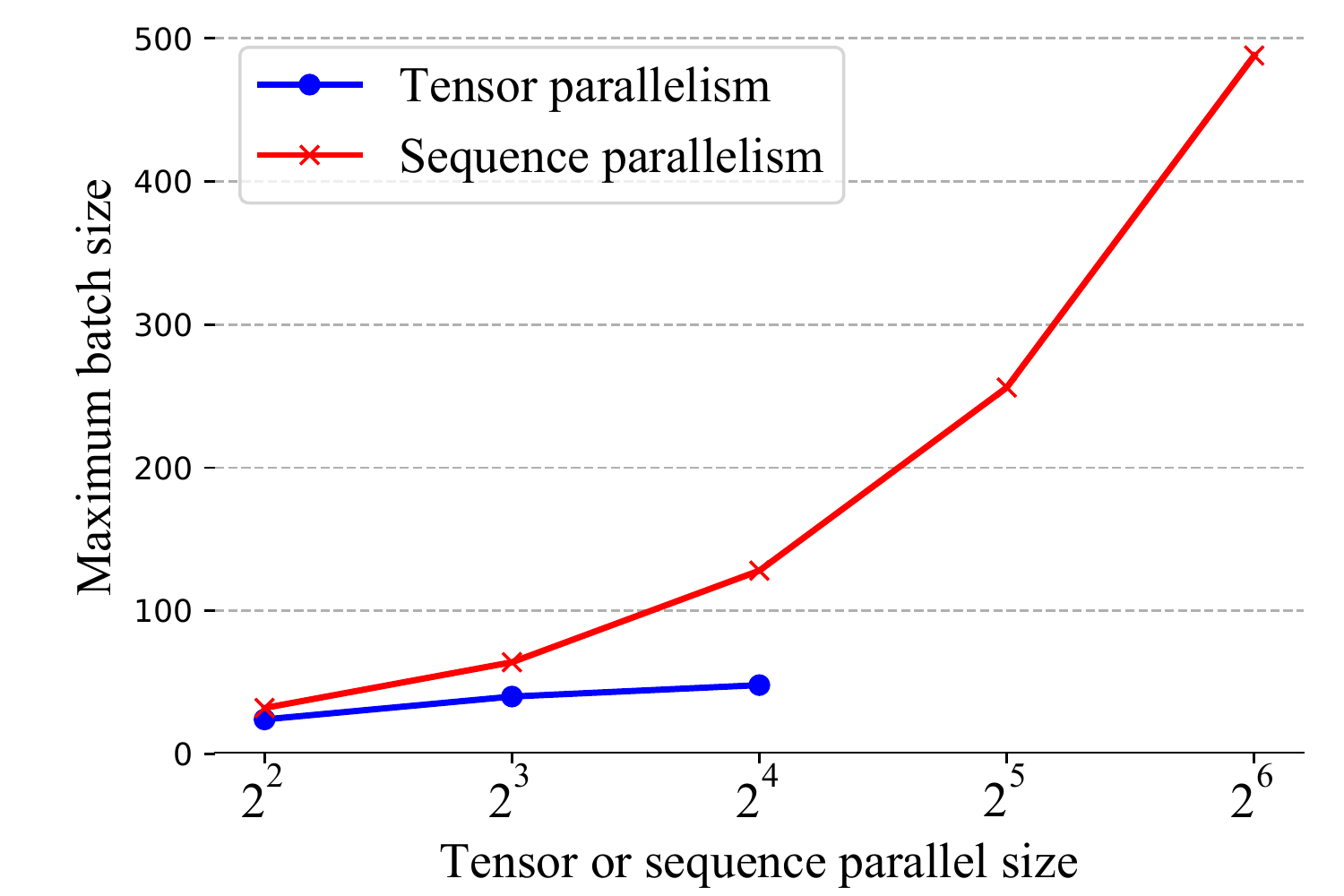}
\subcaption{Maximum batch size of BERT Large scaling along tensor or sequence parallel size}
\label{fig:Batch_BERT_Large_wo_pipeline_batchsize}
\end{minipage}
\begin{minipage}[b]{0.45\linewidth}
\centering
\includegraphics[width=1.0\textwidth]{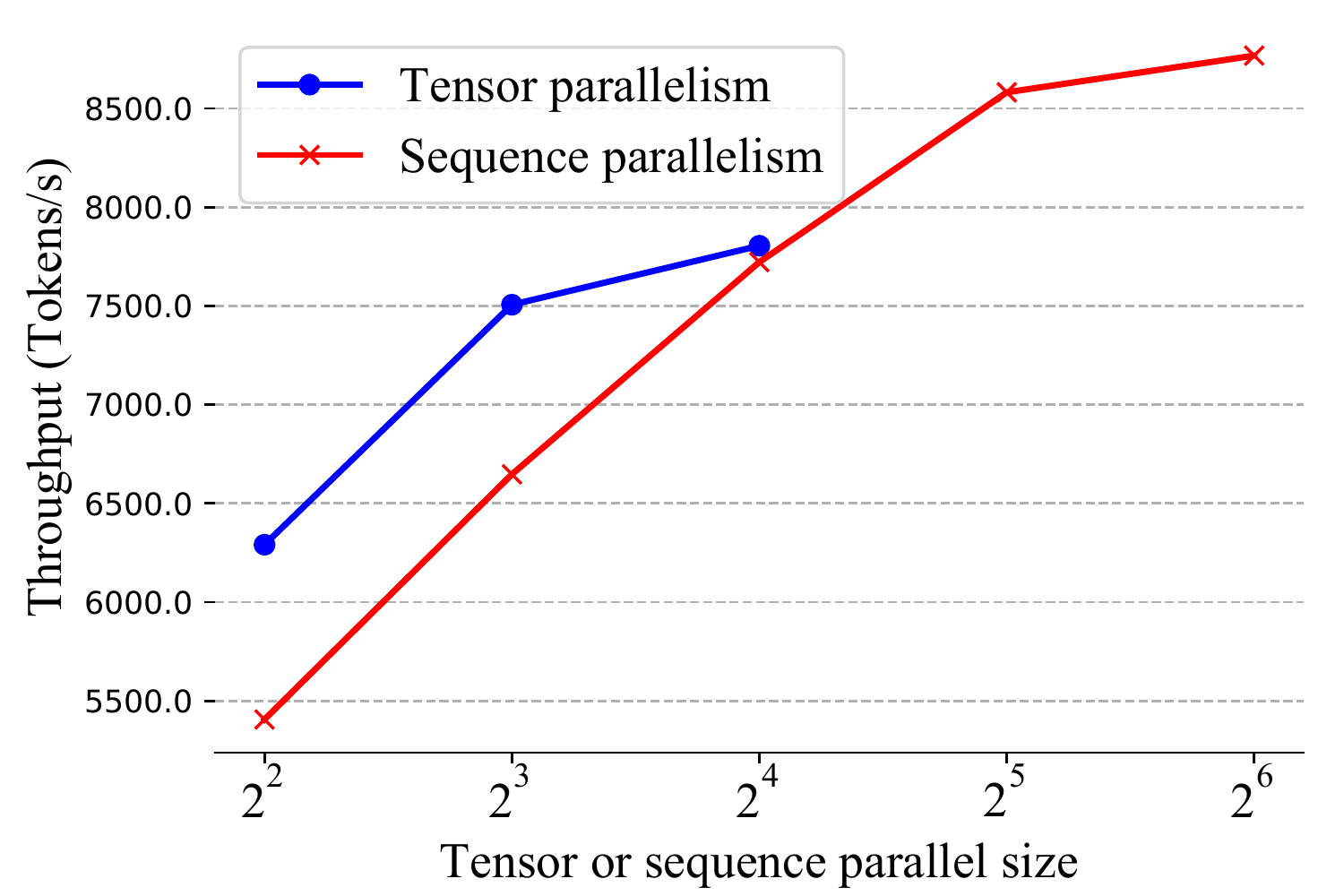}
\subcaption{Throughput of BERT Large scaling along tensor or sequence parallel size}
\label{fig:Batch_BERT_Large_wo_pipeline_throughput}
\end{minipage}
\centering
\caption{Scaling with sequence/tensor parallelism}
\label{fig:maximum batch size strong scaling Large}
\end{figure}

Compared with BERT Base setting, the only difference is, the tensor parallel size is a maximum of 16 for the BERT Large model in Megatron-LM. In Figure~\ref{fig:Batch_BERT_Large_wo_pipeline_batchsize}, our method achieved $2.7$ times larger batch size for BERT Large on 16 GPUs, and the batch size of sequence parallelism on 64 GPUs is $10.2$ times larger than that of tensor parallelism on 16 GPUs. In Figure~\ref{fig:Batch_BERT_Large_wo_pipeline_throughput}, observe that our sequence parallelism achieved comparable throughput with the same parallel size, and more importantly, our system can extend to a larger parallel size to achieve better performance.

\section{Scaling with pipeline parallelism}

\begin{figure}[h]
\centering
\begin{minipage}[b]{0.45\linewidth}
\centering
\includegraphics[width=1.0\textwidth]{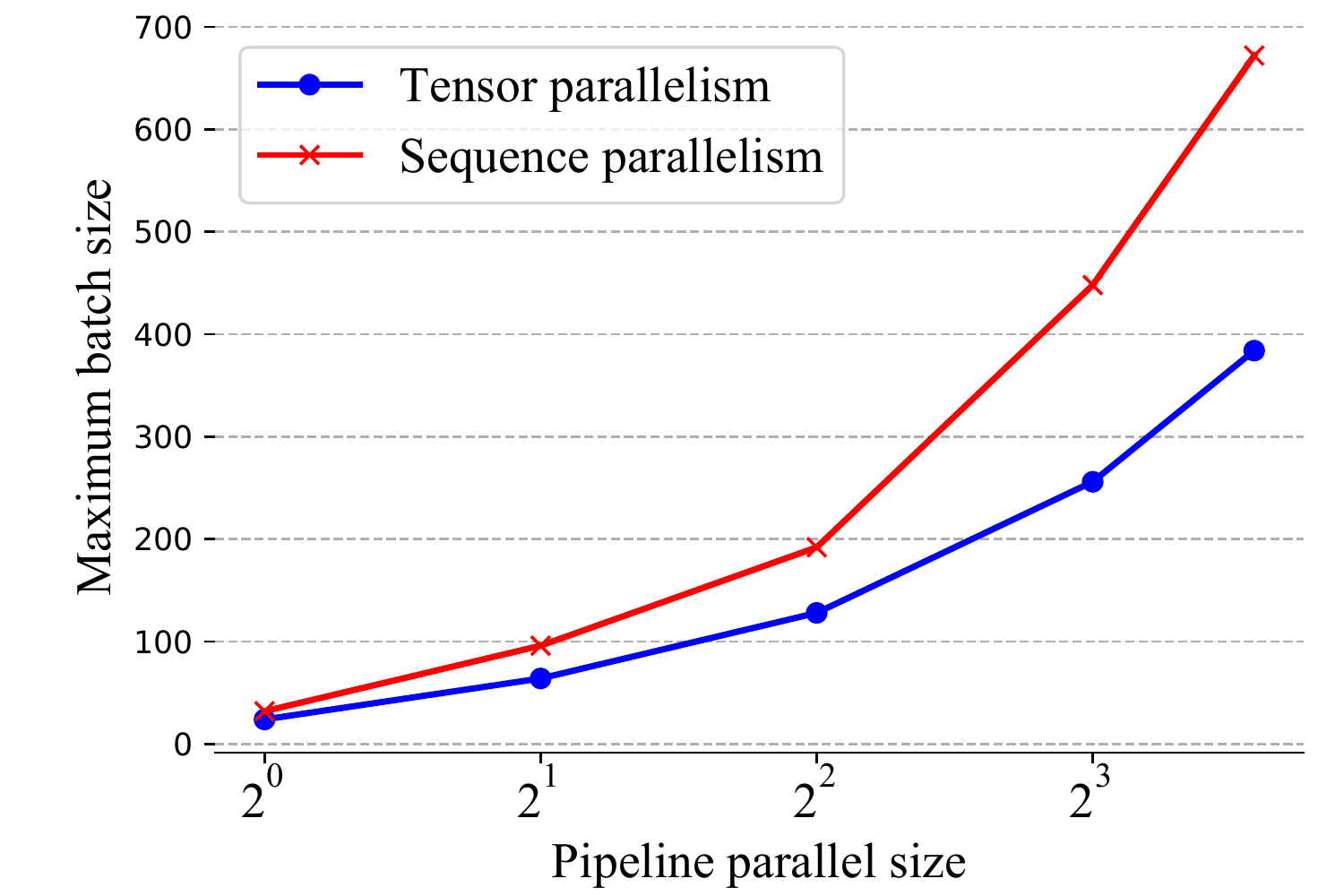}
\subcaption{Maximum batch size of BERT Large scaling along pipeline parallel size}
\label{fig:Batch_BERT_Large_w_pipeline_batchsize}
\end{minipage}
\begin{minipage}[b]{0.45\linewidth}
\centering
\includegraphics[width=1.0\textwidth]{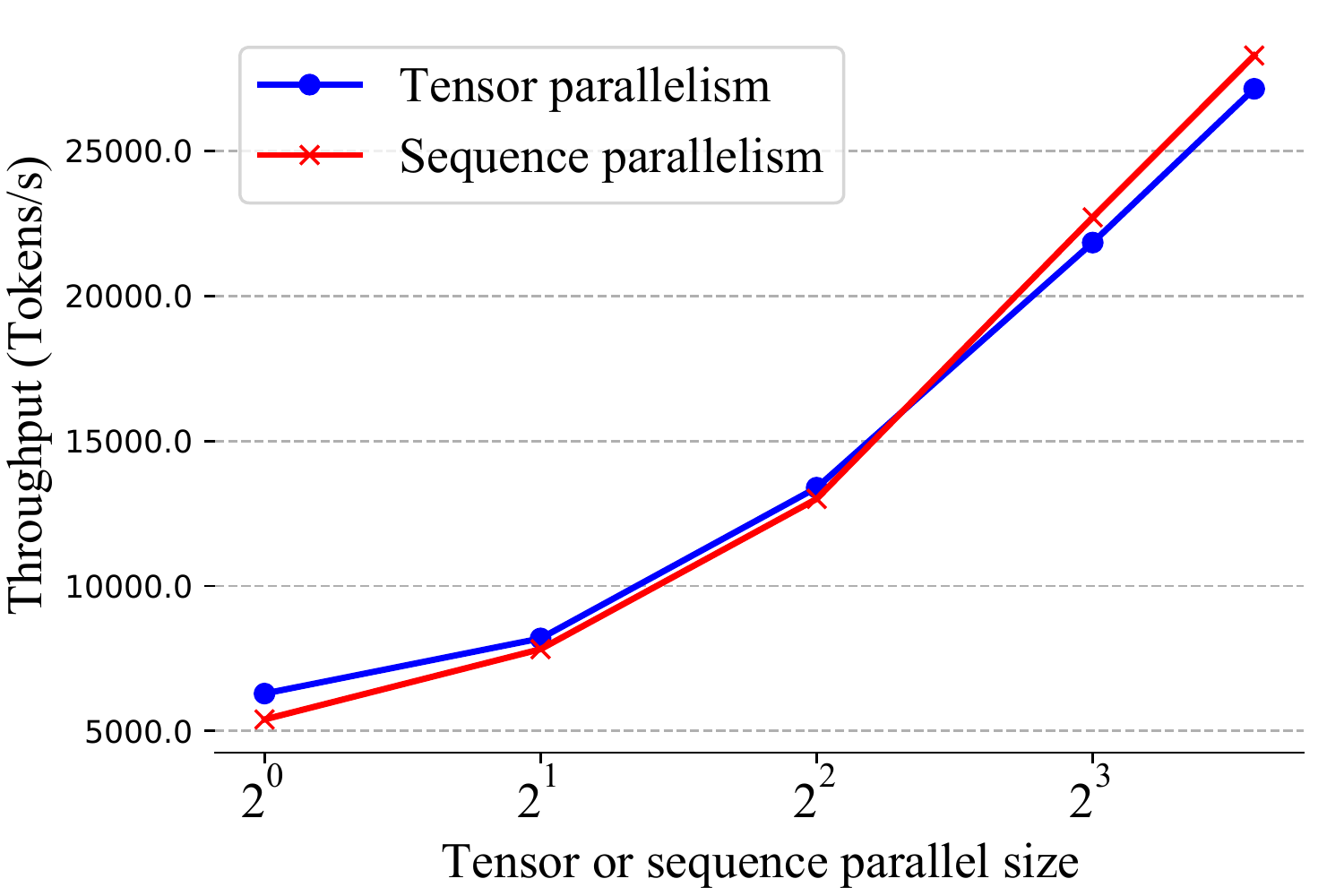}
\subcaption{Throughput of BERT Large scaling along pipeline parallel size}
\label{fig:Batch_BERT_Large_w_pipeline_throughput}
\end{minipage}
\caption{Scaling with pipeline parallelism}
\label{Appendix:fig:maximum batch size weak scaling}
\end{figure}

For BERT Large, sequence parallelism achieved higher maximum batch size than tensor parallelism in Figure~\ref{fig:Batch_BERT_Large_w_pipeline_batchsize}. Sequence parallelism also performs better on throughput when using more pipeline stages as shown in Figure~\ref{fig:Batch_BERT_Large_w_pipeline_throughput}.

\section{Maximum sequence length}\label{Appendix:Maximum sequence length}

\begin{figure}[h]
\centering
\includegraphics[width=0.45\textwidth]{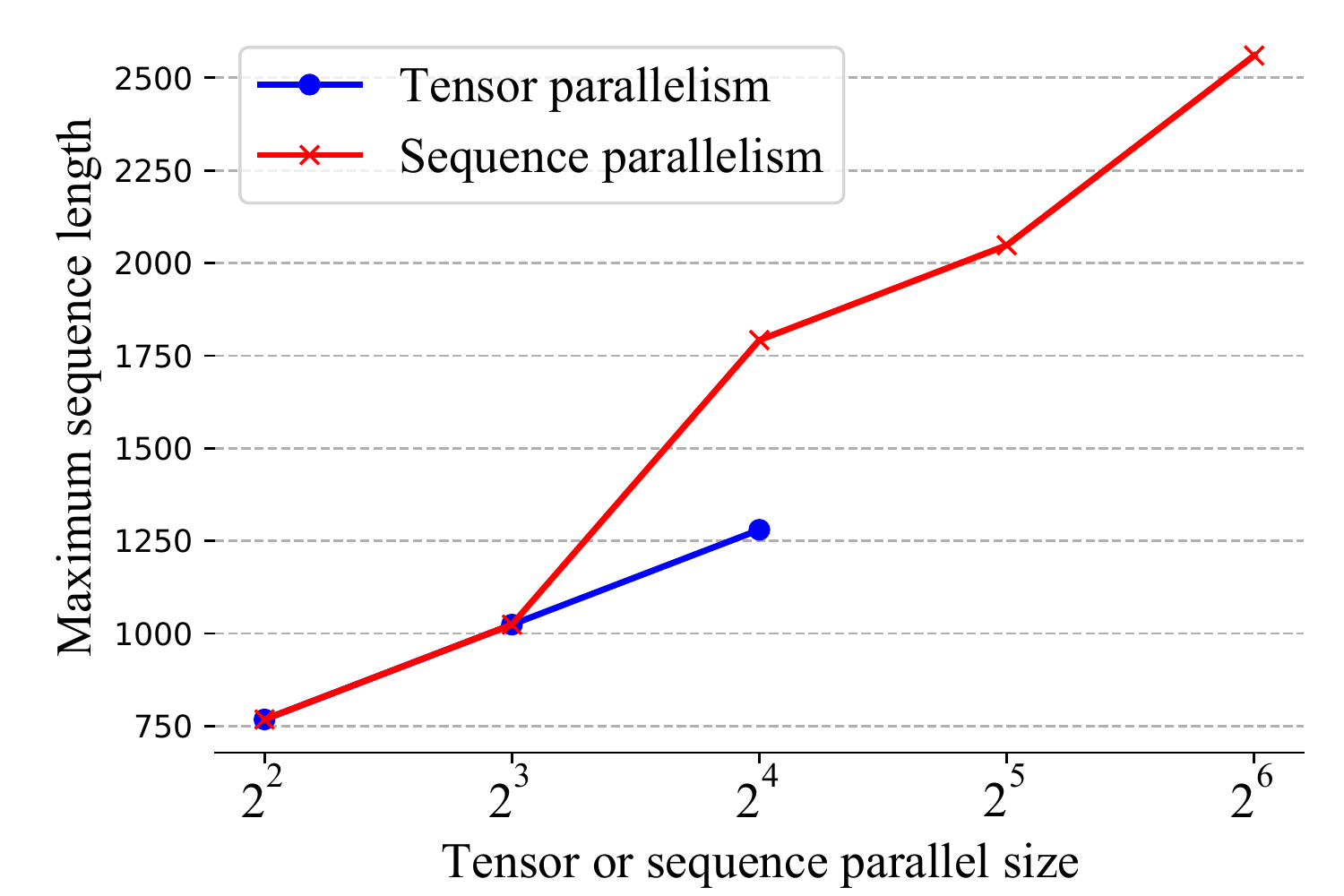}
\caption{Maximum sequence length on BERT Large}
\label{fig:Length_BERT_Large_wo_pipeline_length}

\vspace{-0.3cm}
\end{figure}

\paragraph{BERT Large} Similarly, we compared tensor parallelism without pipeline parallelism. We fixed batch size as 16 for BERT Large and did not use pipeline parallelism. As shown in Figure~\ref{fig:Length_BERT_Large_wo_pipeline_length}. When we scale up to 64 GPUs, we can achieve around $2\times$ maximum sequence length and scale better through splitting a sequence into multiple chunks on BERT Large.



\end{document}